# Self-Organized Operational Neural Networks with Generative Neurons

Serkan Kiranyaz[1], Junaid Malik[1,4], Habib Ben Abdallah[1], Turker Ince[2], Alexandros Iosifidis[3] and Moncef Gabbouj[4]

[1] Electrical Engineering, College of Engineering, Qatar University, Qatar; e-mail: mkiranyaz@qu.edu.qa
[2] Electrical & Electronics Engineering Department, Izmir University of Economics, Turkey; e-mail: turker.ince@ieu.edu.tr
[3] Department of Engineering, Aarhus University, Denmark; e-mail: alexandros.iosifidis@eng.au.dk
[4] Department of Computing Sciences, Tampere University, Finland; e-mail: Moncef.gabbouj@tuni.fi

*Abstract*—**Operational Neural Networks (ONNs) have recently been proposed to address the well-known limitations and drawbacks of conventional Convolutional Neural Networks (CNNs) such as network homogeneity with the sole linear neuron model. ONNs are heterogenous networks with a generalized neuron model that can encapsulate any set of non-linear operators to boost diversity and to learn highly complex and multi-modal functions or spaces with minimal network complexity and training data. However, Greedy Iterative Search (GIS) method, which is the search method used to find optimal operators in ONNs takes many training sessions to find a single operator set per layer. This is not only computationally demanding, but the network heterogeneity is also limited since the same set of operators will then be used for all neurons in each layer. Moreover, the performance of ONNs directly depends on the operator set library used, which introduces a certain risk of performance degradation especially when the optimal operator set required for a particular task is missing from the library. In order to address these issues and achieve an ultimate heterogeneity level to boost the network diversity along with computational efficiency, in this study we propose Self-organized ONNs (Self-ONNs) with generative neurons that have the ability to adapt (optimize) the nodal operator of each connection during the training process. Therefore, Self-ONNs can have an utmost heterogeneity level required by the learning problem at hand. Moreover, this ability voids the need of having a fixed operator set library and the prior operator search within the library in order to find the best possible set of operators. We further formulate the training method to back-propagate the error through the operational layers of Self-ONNs. Experimental results over four challenging problems demonstrate the superior learning capability and computational efficiency of Self-ONNs over conventional ONNs and CNNs, even with more compact networks.**

## I. Introduction

Multi-Layer Perceptrons (MLPs), and their derivatives, Convolutional Neural Networks (CNNs) have a common drawback: they employ a homogenous network structure with identical "linear" neuron model. This naturally makes them only a crude model of the biological neurons or mammalian neural systems, which are heterogeneous and composed of a highly diverse neuron types with distinct biochemical and electrophysiological properties [13]-[18]. With such crude models, conventional homogenous networks can learn sufficiently well problems with a monotonous, relatively simple and linearly separable solution space but they fail to accomplish this whenever the solution space is highly nonlinear and complex [8]-[10], [32], [33]. Despite many attempts to address this deficiency by searching for good network architectures [4], [5] or by following extremely laborious search strategies [6]-[10], or hybrid network models [19]-[21], or new parameter update approaches [22], [23]; no attempts have been made to address the core problem, i.e., the network homogeneity with only linear neurons coming from decades-old McCulloch-Pitts model [11].

To address this drawback, a heterogenous and dense network model, Generalized Operational Perceptrons (GOPs) has recently been proposed [32]-[36]. GOPs aim to model biological neurons with *distinct* synaptic connections. GOPs have demonstrated a superior diversity, encountered in biological neural networks, which resulted in an elegant performance level on numerous challenging problems where conventional MLPs entirely failed [32]-[36] (e.g. Two-Spirals or N-bit parity problems). Following GOPs footsteps, a heterogenous and non-linear network model, called Operational Neural Network (ONN), has recently been proposed [37] as a superset of CNNs. ONNs, like their predecessor GOPs, boost the diversity to learn highly complex and multi-modal functions or spaces with minimal network complexity and training data. More specifically, the diverse set of neurochemical operations in biological neurons (the non-linear synaptic connections plus the integration process occurring in the soma of a biological neuron model) have been modelled by the corresponding "Nodal" (synaptic connection) and "Pool" (integration in soma) operators whilst the "Activation" operator has directly been adopted. A particular set of nodal, pool and activation operator forms an "operator set" and all potential operator sets are stored in an operator set library. Using the so-called Greedy Iterative Search (GIS) method, an optimal operator set per layer can iteratively be searched during several short Back-Propagation (BP) training sessions. The final ONN can then be configured by using the best operator sets found, each of which is assigned to *all* neurons of the corresponding hidden layers.

The results over challenging learning problems demonstrate that 1) with the right operator set, ONNs can perform the required linear or non-linear transformation in each layer/neuron, so as to maximize the learning performance, and, 2) ONNs not only outperforms CNNs significantly, they are even able to learn those problems where CNNs entirely fail. However**,** ONNs proposed in [37], too, exhibit certain drawbacks. First and the foremost is the limited heterogeneity due to the usage of a single operator set for all neurons in a hidden layer. This enforces the *sole* usage of single nodal operator for all kernel connections of each neuron to the neurons in the previous layer. A major limitation is that the learning performance of the ONN directly depends on the operators (particularly nodal operators) in the operator set library, which is fixed in advance. In other words, if the right operator set for a proper learning is missing, the learning performance will deteriorate. Obviously, it is not feasible to cover all possible nodal operators since they are infinitely many. Furthermore, many operators cannot even be formulated with standard non-linear functions, yet, they can be approximated. Finally, the GIS is a computationally demanding local search process which requires many BP runs. The best operator sets found may not be optimal and especially for deep networks that are trained over large-scale datasets, GIS results in a real bottleneck computational complexity.

In order to address these drawbacks and limitations, in this study we propose Self-organized ONNs (Self-ONNs) with generative neurons. Self-ONNs, as the name implies, have the ability to self-organize the network operators during training. Therefore, they neither need any operator set library in advance, nor require any prior search process to find the optimal nodal operator. In fact, the limitation of the usage of a single nodal operator for all kernel connections of each neuron will be addressed by the "generative neurons" where each neuron can create any *combination* of nodal operators, which may not necessarily be a well-defined function such as linear, sinusoids, hyperbolic, exponential or some other standard functions. It is true that the (weights) parameters of the kernel change the nodal operator output, e.g., for a "Sinusoid" nodal operator of a particular neuron, the kernel parameters are distinct frequencies. This allows the creation of "any" harmonic function; however, the final nodal operator function after training cannot have any other pattern or form besides a pure sine wave even though a "composite operator", e.g., the linear combination of harmonics, hyperbolic and polynomial, or an arbitrary nodal operator function would perhaps be a better choice for this neuron than pure sinusoids. This is in fact the case for biological neurons where the synaptic connections can exhibit any arbitrary form or pattern. In brief, a generative-neuron is a neuron with a *composite* nodal-operator that can be generated during training without any restrictions. As a result, with such generative neurons, a Self-ONN can self-organize its nodal operators during training and thus, it will have the nodal operator functions "optimized" by the training process to maximize the learning performance. For instance, in the sample illustration shown in Figure 1, the CNN and ONN neurons have *static* nodal operators (linear and harmonic, respectively) for their 3x3 kernels, while the generative-neuron can have *any* arbitrary nodal function, $\mathbf{\Psi}$, (including possibly standard types such as linear and harmonic functions) for each kernel element of each connection. This is a great flexibility that permits the formation of *any* nodal operator function. Finally, the training method that back-propagates the error through the operational layers of Self-ONNs is formulated in order to generate the right nodal functions of its neurons. Over the same set of challenging problems in [37] with the same severe restrictions, we shall show that Self-ONNs can achieve a comparable and usually better performance levels than the parameter-equivalent ONNs with a superior computational efficiency. The performance gap compared against the equivalent CNNs further widens even for Self-ONNs with significantly fewer neurons and with a short training.

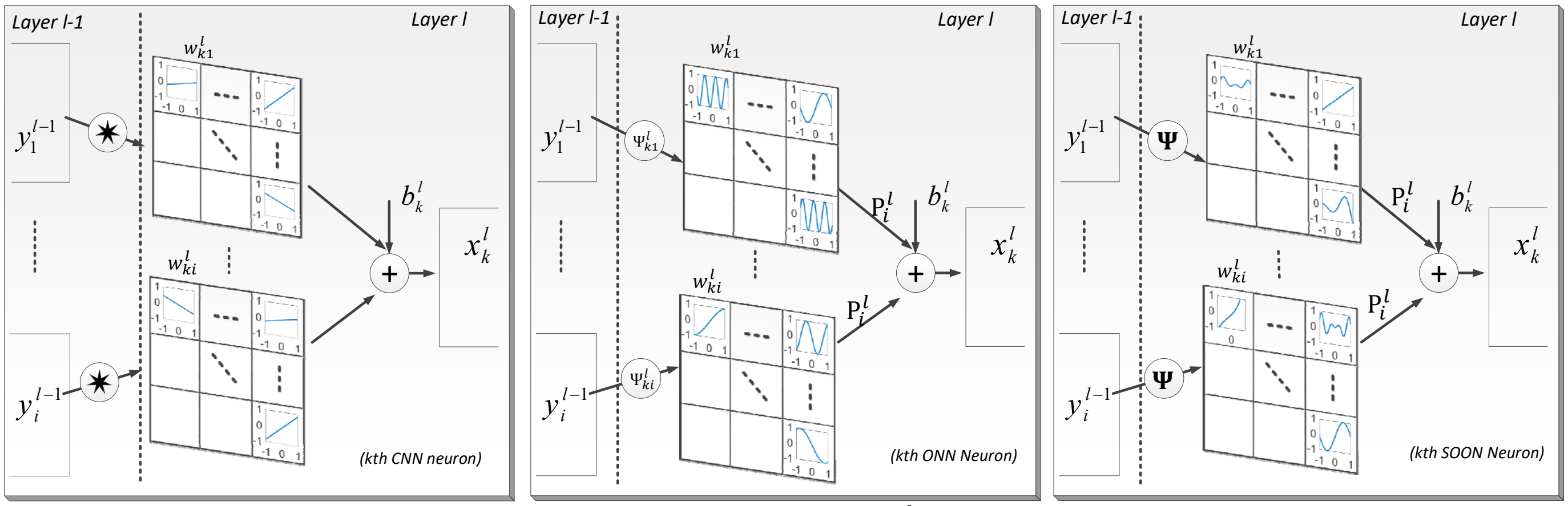

**Figure 1: An illustration of the nodal operations in the kernels of the $k^{th}$ CNN (left), ONN (middle) and Self-ONN (right) neurons at layer *l*.**

The rest of the paper is organized as follows: Section II will briefly present the conventional ONNs while the BP training is summarized in Appendix A. Section III presents Self-ONNs and generative neurons in detail and formulates the forward-propagation (FP) and back-propagation (BP) training. It further discusses major features of Self-ONNs on a toy problem. Section IV presents detailed comparative evaluations among Self-ONNs, ONNs and CNNs over four challenging problems. The computational complexity analysis of these networks for both FP and BP is also presented in this section. Finally, Section V concludes the paper and suggests topics for future research.

## II. Operational Neural Networks

Similar to MLPs, conventional CNNs make use of the classical "linear" neuron model; however, they further apply two restrictions: kernel-wise limited connections and weight sharing. These restrictions turn the linear weighted sum for MLPs to the convolution formula used in CNNs. This is illustrated in Figure 2 (left) where the three consecutive convolutional layers without the sub-sampling (pooling) layers are shown. ONNs borrows the essential idea of GOPs and thus extends the sole usage of linear convolutions in the convolutional neurons by the *nodal* and *pool* operators. This constitute the operational layers and neurons while the two fundamental restrictions, weight sharing and limited (kernel-wise) connectivity, are directly inherited from conventional CNNs. This is also illustrated in Figure 2 (right) where three operational layers and the $k^{\text{th}}$ neuron with 3x3 kernels belong to an ONN. As illustrated, the input map of the $k^{\text{th}}$ neuron at the current layer, $x_k^l$, is obtained by *pooling* the final output maps, $y_i^{l-1}$ of the previous layer neurons *operated* with its corresponding kernels, $w_{ki}^l$, as follows:

$$x_k^l = b_k^l + \sum_{i=1}^{N_{l-1}} oper2D(w_{ki}^l, y_i^{l-1}, 'NoZeroPad')$$

$$x_k^l(m,n)\Big|_{(0,0)}^{(M-1,N-1)} = b_k^l + \quad (1)$$

$$\sum_{i=1}^{N_{l-1}} \left( P_k^l \left[ \begin{matrix} \Psi_{ki}^l\left(w_{ki}^l(0,0), y_i^{l-1}(m,n)\right), \ldots, \\ \Psi_{ki}^l\left(w_{ki}^l(\mathrm{r},\mathrm{t}), y_i^{l-1}(m+r, n+t), \ldots\right), \ldots \end{matrix} \right] \right)$$

A close look to Eq. (1) reveals the fact that when the pool operator is "summation", $P_k^l = \Sigma$, and the nodal operator is "linear", $\Psi_{ki}^l(y_i^{l-1}(m,n), w_{ki}^l(r,t)) = w_{ki}^l(r,t) y_i^{l-1}(m,n)$, for *all* neurons, then the resulting homogenous ONN will be identical to a CNN. Hence, ONNs are indeed a superset of CNNs as the GOPs are a superset of MLPs.

For Back-Propagation (BP) training of an ONN, the following four consecutive stages should be iteratively performed: 1) Computation of the delta error, $\Delta_1^L$, at the output layer, 2) Inter-BP between two consecutive operational layers, 3) Intra-BP in an operational neuron, and 4) Computation of the weight (operator kernel) and bias sensitivities in order to update them at each BP iteration. Stage-3 also takes care of sub-sampling (pooling) operations whenever they are applied in the neuron. BP training is briefly formulated in Appendix A while further details can be obtained from [37].

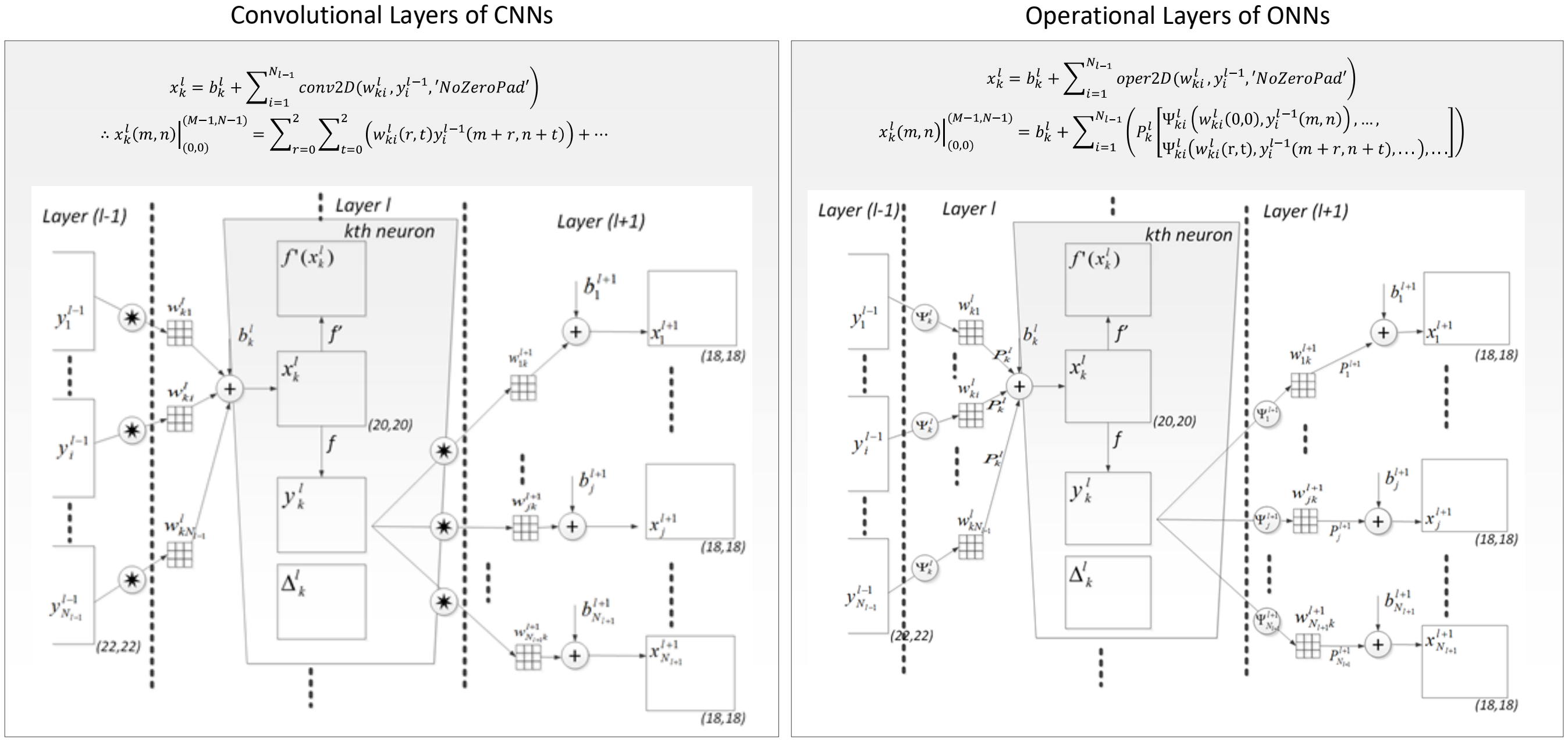

**Figure 2: The illustration of the $k^{\text{th}}$ neuron of a CNN (left) and an ONN (right) along with the three consecutive convolutional (left) and operational (right) layers.**

## III. Self-organized Operational Neural Networks

In this section, first the model of generative neurons which are the main difference between conventional ONNs and Self-ONNs is presented. Then we shall formulate the forward- and back-propagation for Self-ONNs and finally, for the sake of clarity we shall discuss its major characteristics and computational efficiency over a toy problem.

### A. Generative Neurons

As discussed earlier, a generative-neuron is a neuron with a "composite nodal-operator" that is iteratively created during BP training without any restrictions. In this way, each generative-neuron in a Self-ONN can have the self-optimized nodal operators by the BP training for each kernel element and for each connection (to each previous layer neuron) to maximize the learning performance. In order to generate a composite nodal operator, a straightforward choice would be the weighted sum of standard functions. For example, a composite nodal function may have the following expression:

$$\Psi(\mathbf{w}, y) = w_1 Sin(w_2 y) + w_3 exp(w_4 y) + \cdots + w_Q y \quad (2)$$

where $\mathbf{w}$ is a $Q$-dimensional array of parameters that is composed of weights (e.g. $w_1$ and $w_3$ in Eq. (2)) and internal parameters of the individual functions, (e.g., $w_2$ (frequency), $w_4$ (power factor) and $w_Q$ (slope) in Eq. (2)). However, such a straightforward formation of the composite nodal functions would obviously have severe stability issues due to the different dynamic ranges of the individual non-linear functions composed. Moreover, it requires too many parameters to be tuned especially when the operator set library contains many individual nodal operator functions. It is, as well, equally redundant because one can form any arbitrary function by other conventional methods such as Taylor Polynomials or Fourier Series. The former is a better choice due to its lower computational complexity than the latter. The Taylor approximation of a function, $\mathrm{f}(x)$, near a point, $x = a$, can be expressed as,

$$\mathrm{f}(x) = \mathrm{f(a)} + \frac{\mathrm{f}'(\mathrm{a})}{1!}(x-a) + \frac{\mathrm{f}''(\mathrm{a})}{2!}(x-a)^2 + \frac{\mathrm{f}'''(\mathrm{a})}{3!}(x-a)^3 + \cdots \quad (3)$$

where $\mathrm{f}', \mathrm{f}''$ and $\mathrm{f}'''$ are the first, second and third derivatives, respectively. Hence, one can form the composite nodal operator function using the $Q^{th}$ order truncated Taylor approximation as follows:

$$\Psi(\mathbf{w}, \mathrm{y}) = w_0 + w_1(y-a) + w_2(\mathrm{y}-\mathrm{a})^2 + \cdots + w_Q(\mathrm{y}-\mathrm{a})^Q \quad (4)$$

where $w_q = \frac{f^{(n)}(a)}{q!}$ is the $q^{\text{th}}$ parameter of the $Q^{\text{th}}$ order polynomial. The training process optimizes the parameters to form (approximate) the best-fitting nodal operator for each kernel element of each individual inter-neuron connection. An immediate issue arises, this approximation is only valid near the point, $y=a$. The farther the points are from $a$, the coarser the approximation becomes. However, this does not affect Self-ONNs since the nodal operators operate over the neuron outputs of the previous layer, each of which is bounded based on the generative range of the activation operator function. If, for instance, the activation function is a sigmoid, then the outputs, y, operates within the range of [0, 1]. In this case, the nodal operator function can be approximated for $a = 0.5$ (the mid-point) and a sufficiently higher-degree polynomial can approximate any arbitrary function sufficiently well around the close vicinity of this point, i.e., in the range of [0, 1]. In this study, we are using the activation function *tangent hyperbolic* (*tanh*) that is bounded in the range of, [-1, 1]. In this case, naturally, $a = 0$ and the $Q^{th}$ order Taylor approximation in Eq. (4) simplifies to the Maclaurin series as,

$$\Psi(\mathbf{w}, \mathrm{y}) = w_0 + w_1 y + w_2 y^2 + \cdots + w_Q y^Q \quad (5)$$

Finally, the bias coefficient, $w_0$, can be omitted since the overall DC bias will anyway be compensated by each neuron's bias term, $b_i^l$.

### B. Forward Propagation in Self-ONNs

The forward propagation (FP) formula for Self-ONNs differs from FP for ONNs in Eq. (1) by the following two points:

1) Each nodal operator function with the single kernel element, $\Psi_i^{l+1}\left(y_k^l(m+r, n+t), w_{ik}^{l+1}(\mathrm{r},\mathrm{t})\right)$, will now be *approximated* by the composite nodal operator, $\Psi\left(y_k^l(m+r, n+t), \boldsymbol{w}_{ik}^{l+1}(\mathrm{r},\mathrm{t})\right)$, as expressed by the Maclaurin series in Eq. (5),

2) The *scalar* kernel parameter, $w_{ik}^{l+1}(\mathrm{r},\mathrm{t})$, of the kernel of an ONN neuron, is replaced by a $Q$-dimensional array, $\boldsymbol{w}_{ik}^{l+1}(\mathrm{r},\mathrm{t})$, and the Maclaurin series expression in Eq. (5) is the only composite nodal operator function for all neurons in the network. Thus, individual nodal operators, e.g., $\Psi_i^{l+1}$, can now be expressed simply as the composite nodal operator, $\boldsymbol{\Psi}$. So, the composite nodal function for the kernel element, $\boldsymbol{w}_{ik}^{l+1}(\mathrm{r},\mathrm{t})$, can be expressed as follows:

$$\begin{aligned}\boldsymbol{\Psi}\left(y_k^l(m+r, n+t), \boldsymbol{w}_{ik}^{l+1}(\mathrm{r},\mathrm{t})\right) &= w_{ik}^{l+1}(\mathrm{r},\mathrm{t},1) y_k^l(m+r, n+t) \\ &+ w_{ik}^{l+1}(\mathrm{r},\mathrm{t},2) y_k^l(m+r, n+t)^2 \\ &+ \cdots \\ &+ w_{ik}^{l+1}(\mathrm{r},\mathrm{t},\mathrm{Q}) y_k^l(m+r, n+t)^Q\end{aligned} \quad (6)$$

where the DC bias term, $w_{ik}^{l+1}(\mathrm{r},\mathrm{t},0)$, is omitted due to the reasoning mentioned earlier. Therefore, a generative neuron of a Self-ONN has a 3D kernel matrix where the $q^{\text{th}}$ weight of the kernel element $(\mathrm{r},\mathrm{t})$ is represented by $w_{ik}^{l+1}(\mathrm{r},\mathrm{t},\mathrm{q})$. As illustrated in Figure 1, for each neuron in a Self-ONN, any nodal function can be generated (approximated) for each kernel element and for each kernel connection. This results in an enhanced flexibility and diversity even over an ONN neuron where only a standard nodal operator function has to be used for all kernels connected to previous layer neurons. Finally, the generative neurons of a Self-ONN can still have different pool and activation operators; however, in this study we keep the choices fixed to *"summation"* for pool and "*tanh*" for activation.

### C. Back Propagation on Self-ONNs:

For Self-ONNs, the contributions of each pixel in the $M \times N$ output map, $y_k^l(m,n)$ on the next layer input map, $x_i^{l+1}(m,n)$, can now be expressed as in Eq. (7). Using the chain rule, the delta error of the output pixel, $y_k^l(m,n)$, can therefore, be expressed as in Eq. (8) in the generic form of pool, $P_i^{l+1}$, and nodal, $\Psi_i^{l+1}$, operator functions of each operational neuron $i \in [1,..,N_{l+1}]$ in the next layer. In Eq. (8), note that the first term, $\frac{\partial x_i^{l+1}(m-r,n-t)}{\partial P_i^{l+1}[..,\boldsymbol{\Psi}(y_k^l(m,n),\boldsymbol{w}_{ik}^{l+1}(r,t)),..]} = 1$.

Let $\nabla_{\Psi} P_i^{l+1}(m,n,r,t) = \frac{\partial P_i^{l+1}\left[\ldots,\Psi\left(y_k^l(m,n),\boldsymbol{w_{ik}^{l+1}}(\boldsymbol{r},\boldsymbol{t})\right),\ldots\right]}{\partial \Psi\left(y_k^l(m,n),\boldsymbol{w_{ik}^{l+1}}(\boldsymbol{r},\boldsymbol{t})\right)}$ and $\nabla_y \Psi(m,n,r,t) = \frac{\partial \Psi\left(y_k^l(m,n),\boldsymbol{w_{ik}^{l+1}}(\boldsymbol{r},\boldsymbol{t})\right)}{\partial y_k^l(m,n)}$. Then, Eq. (8) simplifies to Eq. (9). Note further that $\Delta y_k^l$, $\nabla_{\Psi_{ki}} P_i^{l+1}$ and $\nabla_y \Psi$ have the same size, $M \times N$ while the next layer delta error, $\Delta_i^{l+1}$, has the size, $(M - K_x + 1) \times (N - K_y + 1)$, respectively. Therefore, to enable this variable 2D convolution in this equation, the delta error, $\Delta_i^{l+1}$, is padded by zeros at all four boundaries ($K_x - 1$ zeros on left and right, $K_y - 1$ zeros on bottom and top). Thus, $\nabla_y \Psi(m,n,r,t)$ can simply be expressed as in Eq. (10).

$$
\begin{aligned}
&x_i^{l+1}(m-1,n-1) = \ldots + P_i^{l+1}\left[\Psi\left(y_k^l(m-1,n-1),\boldsymbol{w_{ik}^{l+1}}(\boldsymbol{0},\boldsymbol{0})\right),\ldots,\Psi\left(y_k^l(m,n),\boldsymbol{w_{ik}^{l+1}}(\boldsymbol{1},\boldsymbol{1})\right)\right] + \ldots \\
&x_i^{l+1}(m-1,n) = \ldots + P_i^{l+1}\left[\Psi\left(y_k^l(m-1,n),\boldsymbol{w_{ik}^{l+1}}(\boldsymbol{0},\boldsymbol{0})\right),\ldots,\Psi\left(y_k^l(m,n),\boldsymbol{w_{ik}^{l+1}}(\boldsymbol{1},\boldsymbol{0})\right),\ldots\right] + \ldots \\
&\boldsymbol{x_i^{l+1}(m,n)} = \ldots + P_i^{l+1}\left[\Psi\left(y_k^l(m,n),\boldsymbol{w_{ik}^{l+1}}(\boldsymbol{0},\boldsymbol{0})\right),\ldots,\Psi\left(y_k^l(m+r,n+t),\boldsymbol{w_{ik}^{l+1}}(\mathbf{r},\mathbf{t}),\right)\ldots\right)\right] + \ldots \\
&\ldots\ldots \\
&\therefore\ x_i^{l+1}(m-r,n-t)\Big|_{(1,1)}^{(M-1,N-1)} = b_i^{l+1} + \sum_{k=1}^{N_1} P_i^{l+1}\left[\ldots,\Psi\left(y_k^l(m,n),\boldsymbol{w_{ik}^{l+1}}(\boldsymbol{r},\boldsymbol{t})\right),\ldots\right]
\end{aligned}
\tag{7}
$$

$$
\begin{aligned}
&\therefore\ \frac{\partial E}{\partial y_k^l}(m,n)\Bigg|_{(0,0)}^{(M-1,N-1)} = \Delta y_k^l(m,n) = \\
&\sum_{i=1}^{N_{l+1}}\left(\sum_{r=0}^{Kx-1}\sum_{t=0}^{Ky-1} \frac{\partial E}{\partial x_i^{l+1}(m-r,n-t)} \times \frac{\partial x_i^{l+1}(m-r,n-t)}{\partial P_i^{l+1}\left[\ldots,\Psi\left(y_k^l(m,n),\boldsymbol{w_{ik}^{l+1}}(\boldsymbol{r},\boldsymbol{t})\right),\ldots\right]} \times \frac{\partial P_i^{l+1}\left[\ldots,\Psi\left(y_k^l(m,n),\boldsymbol{w_{ik}^{l+1}}(\boldsymbol{r},\boldsymbol{t})\right),\ldots\right]}{\partial \Psi\left(y_k^l(m,n),\boldsymbol{w_{ik}^{l+1}}(\boldsymbol{r},\boldsymbol{t})\right)} \times \frac{\partial \Psi\left(y_k^l(m,n),\boldsymbol{w_{ik}^{l+1}}(\mathbf{r},\boldsymbol{t})\right)}{\partial y_k^l(m,n)}\right)
\end{aligned}
\tag{8}
$$

$$
\begin{aligned}
&\Delta y_k^l(m,n)\Big|_{(0,0)}^{(M-1,N-1)} = \\
&\sum_{i=1}^{N_{l+1}}\left(\sum_{r=0}^{K_x-1}\sum_{t=0}^{K_y-1} \Delta_i^{l+1}(m-r,n-t) \times \nabla_{\Psi} P_i^{l+1}(m,n,r,t) \times \nabla_y \Psi(m,n,r,t)\right) \\
&Let\ \nabla_y P_i^{l+1}(m,n,r,t) = \nabla_{\Psi} P_i^{l+1}(m,n,r,t) \times \nabla_y \Psi(m,n,r,t), then \\
&\Delta y_k^l = \sum_{i=1}^{N_{l+1}} Conv2Dvar\left\{\Delta_i^{l+1}, \nabla_y P_i^{l+1}(m,n,r,t)\right\}
\end{aligned}
\tag{9}
$$

$$
\nabla_y \Psi(m,n,r,t) = w_{ik}^{l+1}(\mathrm{r},\mathrm{t},1) + 2w_{ik}^{l+1}(\mathrm{r},\mathrm{t},2)y_k^l(m,n) + \cdots + Qw_{ik}^{l+1}(\mathrm{r},\mathrm{t},\mathrm{Q})y_k^l(m,n)^{Q-1} \tag{10}
$$

Eq. (9) is similar to the corresponding one for ONNs in Eq. (26) in the Appendix, except that there is no need to register a 4D matrix for $\nabla_y \Psi$ since it can directly be computed by Eq. (10). Moreover, when the pool operator is *sum*, $P_i^{l+1} = \Sigma$, then $\nabla_{\Psi} P_i^{l+1}(m,n,r,t) = 1$ and thus, $\nabla_y P_i^{l+1}(m,n,r,t) = \nabla_y \Psi(m,n,r,t)$ which is expressed in Eq. (10).

Once the $\Delta y_k^l$ is computed, using the chain-rule, one can express,

$$
\Delta_k^l = \frac{\partial E}{\partial x_k^l} = \frac{\partial E}{\partial y_k^l}\frac{\partial y_k^l}{\partial x_k^l} = \frac{\partial E}{\partial y_k^l} f'(x_k^l) = \Delta y_k^l f'(x_k^l) \tag{11}
$$

When there is a down-sampling by factors, *ssx* and *ssy*, then the back-propagated delta-error by Eq. (26) should be first up-sampled to compute the delta-error of the neuron. Let zero order up-sampled map be: $uy_k^l = \underset{ssx,ssy}{\mathrm{up}}(y_k^l)$. Then Eq. (11) can be modified, as follows:

$$
\begin{aligned}
\Delta_k^l &= \frac{\partial E}{\partial x_k^l} = \frac{\partial E}{\partial y_k^l}\frac{\partial y_k^l}{\partial x_k^l} = \frac{\partial E}{\partial y_k^l}\frac{\partial y_k^l}{\partial uy_k^l}\frac{\partial uy_k^l}{\partial x_k^l} \\
&= \underset{ssx,ssy}{\mathrm{up}}(\Delta y_k^l)\beta f'(x_k^l)
\end{aligned}
\tag{12}
$$

where $\beta = \frac{1}{ssx.ssy}$ since each pixel of $y_k^l$ is now obtained by averaging ($ssx.ssy$) number of pixels of the intermediate output, $uy_k^l$. Finally, when there is a up-sampling by factors, *usx* and *usy*, then let the average-pooled map be: $dy_k^l = \underset{usx,usy}{\mathrm{down}}(y_k^l)$. Then Eq. (27) can be updated as follows:

$$\Delta_k^l = \frac{\partial E}{\partial x_k^l} = \frac{\partial E}{\partial y_k^l}\frac{\partial y_k^l}{\partial x_k^l} = \frac{\partial E}{\partial y_k^l}\frac{\partial y_k^l}{\partial dy_k^l}\frac{\partial dy_k^l}{\partial x_k^l}$$
$$= \underset{usx,usy}{\text{down}}(\Delta y_k^l)\beta^{-1} f'(x_k^l) \quad (13)$$

*D. Computation of the Weight (Kernel) and Bias Sensitivities*

Recall the expression between an individual kernel weight array, $\boldsymbol{w_{ik}^{l+1}}(\mathbf{r},\mathbf{t})$, and the input map of the next layer, $x_i^{l+1}(m,n)$:

$$x_i^{l+1}(m,n)\Big|_{(1,1)}^{(M-1,N-1)} = b_i^{l+1} + \sum_{i=1}^{N_{l-1}} P_i^{l+1}\left[\begin{matrix}\boldsymbol{\Psi}\left(y_k^l(m,n),\boldsymbol{w_{ik}^{l+1}}(\mathbf{0},\mathbf{0})\right),..,\\ \boldsymbol{\Psi}\left(y_k^l(m+r,n+t),\boldsymbol{w_{ik}^{l+1}}(\mathbf{r},\mathbf{t})\right)...\end{matrix}\right] \quad (14)$$

where the $q^{\text{th}}$ element of the array, $\boldsymbol{w_{ik}^{l+1}}(\mathbf{r},\mathbf{t})$, contributes to all the pixels of $x_i^{l+1}(m,n)$ as expressed in Eq. (6). By using the chain rule of partial derivatives, one can express the weight sensitivities, $\frac{\partial E}{\partial w_{ik}^{l+1}}$, in Eq. (15). A close look to Eq. (6) reveals that, $\frac{\partial \boldsymbol{\Psi}\left(y_k^l(m+r,n+t),\boldsymbol{w_{ik}^{l+1}}(\boldsymbol{r},\boldsymbol{t})\right)}{\partial w_{ik}^{l+1}(r,t,q)} = y_k^l(m+r,n+t)^q$, which then simplifies to Eq. (16) Note that in this equation, the first term, $\Delta_{\text{I}}^{l+1}(m,n)$, is independent from the kernel indices, *r* and *t*. It will be element-wise multiplied by the other two latter terms, each with the same dimension $(\text{M}-\text{Kx}+1)x(N-\text{Ky}+1)$, and created by derivative functions of nodal and pool operators applied over the shifted pixels of $y_k^l(m+r,n+t)$ and the corresponding weight value, $\boldsymbol{w_{ik}^{l+1}}(\boldsymbol{r},\boldsymbol{t})$.

$$\frac{\partial E}{\partial w_{ik}^{l+1}}(r,t,q)\Bigg|_{(0,0,1)}^{(\text{Kx}-1,\text{Ky}-1,\text{Q})} = \sum_{m=0}^{M-\text{Kx}+1}\sum_{n=0}^{N-\text{Ky}+1} \frac{\partial E}{\partial x_{\text{I}}^{l+1}(m,n)} \times \frac{\partial x_{\text{I}}^{l+1}(m,n)}{\partial P_i^{l+1}\left[\boldsymbol{\Psi}\left(y_k^l(m,n),\boldsymbol{w_{ik}^{l+1}}(\mathbf{0},\mathbf{0})\right),..,\boldsymbol{\Psi}\left(y_k^l(m+r,n+t),\boldsymbol{w_{ik}^{l+1}}(\mathbf{r},\mathbf{t})\right)...\right]} \times \frac{\partial P_i^{l+1}\left[\boldsymbol{\Psi}\left(y_k^l(m,n),\boldsymbol{w_{ik}^{l+1}}(\mathbf{0},\mathbf{0})\right),..,\boldsymbol{\Psi}\left(y_k^l(m+r,n+t),\boldsymbol{w_{ik}^{l+1}}(\mathbf{r},\mathbf{t})\right)...\right]}{\partial \boldsymbol{\Psi}\left(y_k^l(m+r,n+t),\boldsymbol{w_{ik}^{l+1}}(\boldsymbol{r},\boldsymbol{t})\right)} \times \frac{\partial \boldsymbol{\Psi}\left(y_k^l(m+r,n+t),\boldsymbol{w_{ik}^{l+1}}(\boldsymbol{r},\boldsymbol{t})\right)}{\partial w_{ik}^{l+1}(r,t,q)} \quad (15)$$

$$\textit{where}\ \frac{\partial x_{\text{I}}^{l+1}(m,n)}{\partial P_i^{l+1}\left[\boldsymbol{\Psi}\left(y_k^l(m,n),\boldsymbol{w_{ik}^{l+1}}(\mathbf{0},\mathbf{0})\right),..,\boldsymbol{\Psi}\left(y_k^l(m+r,n+t),\boldsymbol{w_{ik}^{l+1}}(\mathbf{r},\mathbf{t})\right)...\right]} = 1\ \textit{and}\ \frac{\partial \boldsymbol{\Psi}\left(y_k^l(m+r,n+t),\boldsymbol{w_{ik}^{l+1}}(\boldsymbol{r},\boldsymbol{t})\right)}{\partial w_{ik}^{l+1}(r,t,q)} = y_k^l(m+r,n+t)^q$$

$$\therefore \frac{\partial E}{\partial w_{ik}^{l+1}}(r,t,q)\Bigg|_{(0,0,1)}^{(\text{Kx}-1,\text{Ky}-1,\text{Q})} = \sum_{m=0}^{M-\text{Kx}}\sum_{n=0}^{N-\text{Ky}} \Delta_{\text{I}}^{l+1}(m,n) \times \nabla_{\boldsymbol{\Psi}}\text{P}_i^{l+1}(m+r,n+t,r,t) \times y_k^l(m+r,n+t)^q \quad (16)$$

*If* $\text{P}_\text{i}^{\text{l}+1} = \Sigma$, *then*

$$\frac{\partial E}{\partial w_{ik}^{l+1}}(r,t,q)\Bigg|_{(0,0,1)}^{(\text{Kx}-1,\text{Ky}-1,\text{Q})} = \sum_{m=0}^{M-\text{Kx}}\sum_{n=0}^{N-\text{Ky}} \Delta_{\text{I}}^{l+1}(m,n) \times y_k^l(m+r,n+t)^q$$
$$\therefore \frac{\partial E}{\partial w_{ik}^{l+1}}\langle q\rangle = conv2D\left(\Delta_i^{l+1},\left(\text{y}_\text{k}^\text{l}\right)^\text{q},'NoZeroPad'\right) \quad (17)$$

$$\frac{\partial E}{\partial b_k^l} = \sum_m\sum_n \frac{\partial E}{\partial x_k^l(m,n)}\frac{\partial x_k^l(m,n)}{\partial b_k^l} = \sum_m\sum_n \Delta_k^l\,(m,n) \quad (18)$$

Eq. (16) is somewhat similar to Eq. (33), the corresponding one for conventional ONNs, except that there is no need to register a 4D matrix for $\nabla_w \boldsymbol{\Psi} = y_k^l(m+r,n+t)^q$ since it can directly be computed from the outputs of the neurons. Moreover, when the pool operator is the *sum*, then $\nabla_{\boldsymbol{\Psi}} \mathrm{P}_i^{l+1}(m,n,r,t) = 1$ and Eq. (16) will simplify to Eq. (17) where $\frac{\partial E}{\partial w_{ik}^{l+1}}\langle q \rangle$ is the $q^{th}$ 2D sensitivity kernel, which contains the updates (SGD sensitivities) for the weights of the $q^{th}$ order outputs in Maclaurin polynomial. Finally, the bias sensitivity expressed in Eq. (18) is the same for ONNs and CNNs since the bias is the common additive term for all.

**Algorithm 1: Back-Propagation algorithm for Self-ONNs**

**Input**: Self-*ONN*, *Stopping Criteria* $(maxIter, minMSE)$
**Output**: Self-*ONN** = *BP(Self-ONN*, $iterMax, minMSE)$

1) Initialize network parameters randomly (i.e., ~U(-a, a))
2) UNTIL a stopping criterion is reached, ITERATE:
   a. For each mini-batch in the train dataset, DO:
      i. FP: Forward propagate from the input layer to the output layer to find $q^{th}$ **order outputs**, $(y_k^l)^q$ and the required derivatives and sensitivities for BP such as $f'(x_k^l)$, $\nabla_y \Psi_{ki}^{l+1}$, $\nabla_{\Psi_{ki}} \mathrm{P}_i^{l+1}$ and $\nabla_w \Psi_{ki}^{l+1}$ of each neuron,*k*, at each layer, *l*.
      ii. BP: Using Eq. (23) compute delta error at the output layer and then using Eqs. **(10)** and **(12)**, back-propagate the error back to the first hidden layer to compute delta errors of each neuron, *k*, $\Delta_k^l$ at each layer, *l*.
      iii. PP: Find the bias and weight sensitivities using Eqs. **(17)** and **(18)**, respectively.
      *iv.* Update: Update the weights and biases with the (cumulation of) sensitivities found in previous step scaled with the learning factor, ε, as in Eq. (19):
3) Return Self-*ONN**

Let $w_{ik}^{l+1}\langle q \rangle$ be the $q^{th}$ 2D sub-kernel where $q$=1..$Q$ and composed of *kernel elements*, $w_{ik}^{l+1}(\mathrm{r,t,q})$. During each BP iteration, $t$, the kernel parameters (weights), $w_{ik}^{l+1}\langle q \rangle(t)$, and biases, $b_i^l(t)$, of each neuron in the Self-ONN are updated until a stopping criterion is met. Let, ε(t), be the learning factor at iteration, *t*. One can express the update for the weight kernel and bias at each neuron, *i*, at layer, *l* as follows:

$$w_{ik}^{l+1}\langle q \rangle(t+1) = w_{ik}^{l+1}\langle q \rangle(t) - \varepsilon(t)\frac{\partial E}{\partial w_{ik}^{l+1}}\langle q \rangle$$
$$b_i^l(t+1) = b_i^l(t) - \varepsilon(t)\frac{\partial E}{\partial b_i^l} \qquad (19)$$

As a result, the pseudo-code for BP is presented in Alg. 1.

### *E. Discussions*

Recall that the main difference between ONNs and Self-ONNs is the presence of generative neurons with the composite nodal operator, which is a $Q^{th}$ order Maclaurin polynomial. As a result, each kernel element is a $Q$-dimensional array and therefore, the weight kernels, $w_{ik}^l$, are 3D matrices that are equivalent to an array of Q 2D matrices, $w_{ik}^{l+1}\langle q \rangle, q = 1,..,Q$. Naturally, the weight sensitivities, $\frac{\partial E}{\partial w_{ik}^{l+1}}$, are 3D matrices too. In order to speed-up both FP and BP, the $q^{th}$ power of the neuron outputs, $y_k^l(m,n)^q$, can be computed *only* once (during FP) and stored in individual 3D matrices to be used repeatedly during BP. This is a memory overhead of Self-ONNs compared to ONNs. On the other hand, Self-ONNs do not need the 4D matrices, $\nabla_y \Psi_{ki}^{l+1}$, and $\nabla_w \Psi_{ki}^{l+1}$, both of which can be computed directly.

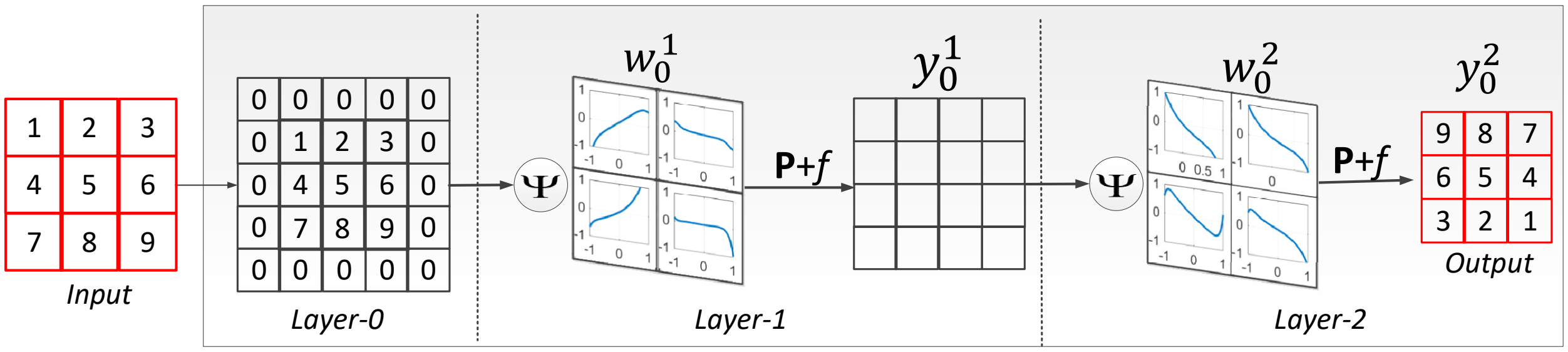

**Figure 3: A sample 3-layer Self-ONN network trained for the toy problem, "rotate 180°".**

For visualization, a Self-ONN with a single hidden layer and a single neuron is trained by BP over the toy problem shown in Figure 3. Input and output are both 3x3 images and the sample Self-ONN has a single input, hidden and output neuron with 2x2 kernels. The toy problem is to learn (regress) to rotate 180° the input image. The final 13th order nodal operators generated during BP are shown in Figure 3, plotted for each kernel element. It is interesting to see optimized nodal operators resembling to a sinusoid, exponential and logarithm with certain variations. This simple Self-ONN network can achieve ~30 times less MSE (or ~14dB higher SNR) than the equivalent CNN trained under the same BP hyperparameters.

## IV. Experimental Results

The comparative evaluations are performed with the same experimental setup and over the same challenging problems in [37]: 1) Image Synthesis, 2) Denoising, 3) Face Segmentation, and 4) Image Transformation with the same training constraints:

i) Low Resolution: 60x60 pixels,
ii) Compact/Shallow Models: *Inx16x32xOut* (for CNN and ONN) and *Inx6x10xOut* for Self-ONNs,
iii) Scarce Train Data: 10% of the dataset
iv) Multiple Regressions per network,
v) Shallow Training: 240 iterations.

For a fair evaluation, we have used a Self-ONN configuration, *Inx6x10xOut* with $Q = 7$ in all layers. In this way all networks have approximately the same number of network parameters. Note that this equivalence results in Self-ONNs having three times less number of hidden neurons than CNNs and ONNs, i.e., 16 vs. 48. Moreover, as in [37] the first hidden layer applies sub-sampling by $ssx = ssy = 2$, and the second one applies up-sampling by $usx = usy = 2$. Self-ONNs are trained using Stochastic Gradient Descent (SGD) without momentum but with a fixed learning parameter whereas adaptive learning rate was applied for CNNs and ONNs in [37]. Finally, three BP runs have also been performed for Self-ONNs and the Self-ONN model that achieved the minimum loss (MSE) during these runs is used for each problem.

### *A. Learning Performance Evaluations*

For each problem, the results obtained by Self-ONNs are compared against the best results obtained by the CNN and ONN. In order to evaluate the learning performance for the regression problems, image denoising, syntheses and transformation, we used the Signal-to-Noise Ratio (SNR) evaluation metric, which is defined as the ratio of the signal power to noise power, i.e., $SNR = 10\log\left(\sigma_{signal}^2/\sigma_{noise}^2\right)$. The ground-truth image is the original signal and its difference to the actual output yields the "noise" image. For the (face) segmentation problem we used the conventional evaluation metrics such as classification error (*CE*) and *F1-score*. For Image Synthesis and Denoising, the benchmark datasets are partitioned into train (10%) and test (90%) for 10-fold cross validation. So, for each fold, all network types are trained 10 times by BP over the train partition and tested over the rest. The following sub-sections will now present the results and comparative evaluations of each problem by the proposed Self-ONNs, ONNs and CNNs.

#### *1) Image Denoising*

As in [37] gray-scale 1500 images from Pascal VOC database are down-sampled and used as the target outputs while the images corrupted by and Gaussian White Noise (GWN) are the input with SNR = 0dB. Compared to earlier denoising works using deep CNNs [37]-[41], this task is far more challenging due to the severity of the noise level applied (0 dB) while all other studies the "noisy" images have SNR levels higher than 15dB. Moreover, the aforementioned restrictions enforce severe training constraints, thus making the problem even more challenging for any machine learning approach.

Figure 4 shows SNR plots of the best denoising results of the three networks for 10 folds and over both partitions. In both train and test partitions, Self-ONNs achieve significantly higher performance as compared to CNNs and ONNs. This is despite the fact that it has three times less neurons. The average SNR levels of CNNs, ONNs and Self-ONNs denoising for the (train) and (test) partitions are: (5.67dB, 5.68dB and 7.05dB), and (5.61dB, 5.46dB and 6.15dB), respectively. Therefore, Self-ONNs can achieve higher than 0.5dB SNR level on the average on the test partition.

Figure 5 presents the SNR vs. iteration plots of all networks for the 1st fold. The convergence speed of Self-ONN can easily be distinguished here, i.e., in both train and test partitions, within only 11 iterations it can reach up the maximum SNR levels of both CNNs and ONNs. This basically shows the crucial role of the optimized nodal operators of its generative neurons. In other words, those "custom-made" nodal operators can quickly be "tuned" within few BP iterations to achieve a superior generalization ability of the network. In this problem Self-ONN has already achieved above 6dB SNR level on the test set in less than 50 iterations.

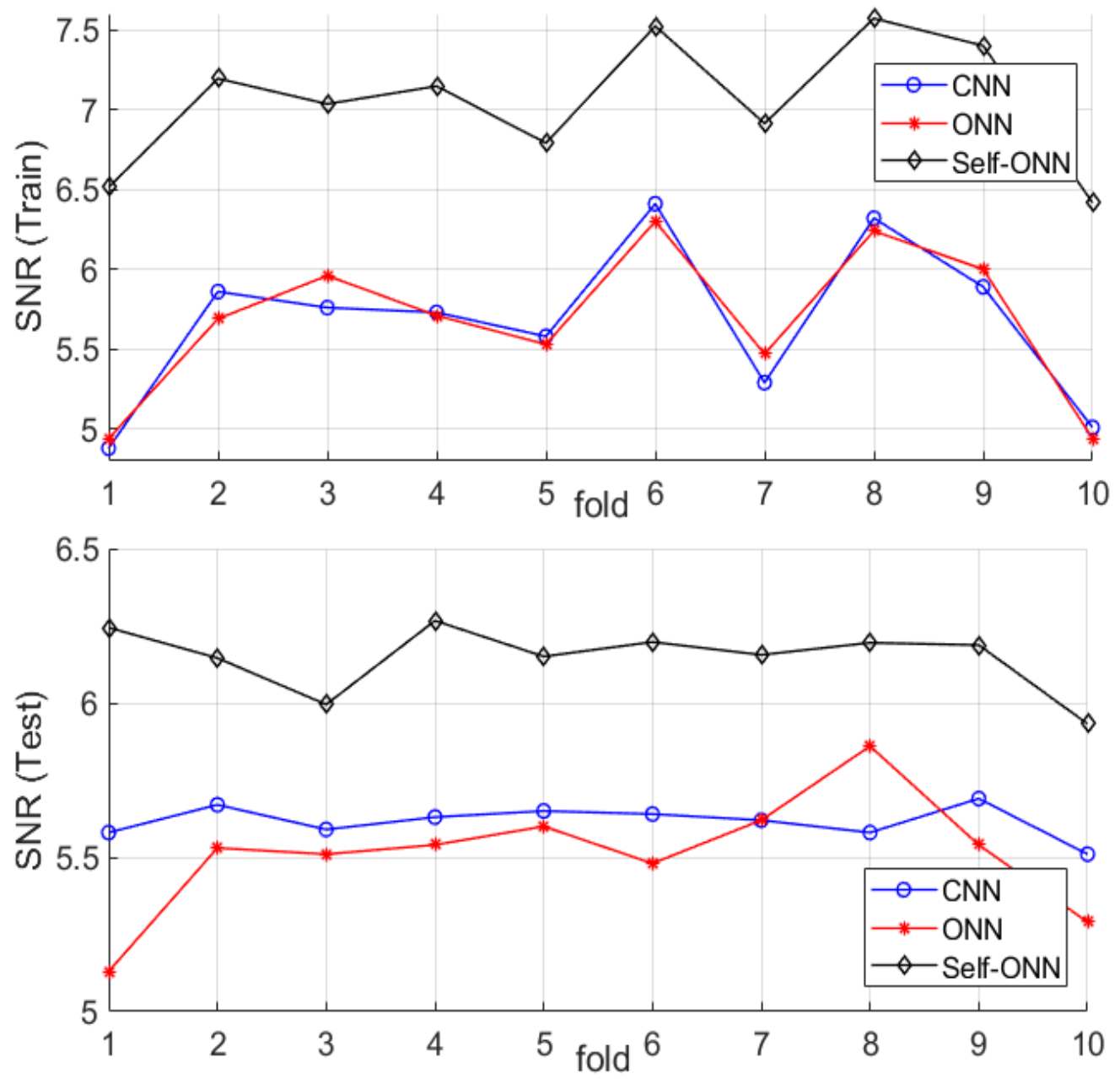

**Figure 4: Best denoising SNR levels for each fold achieved by Self-ONN (black), CNNs (blue) and ONNs (red) in train (top) and test (bottom) partitions.**

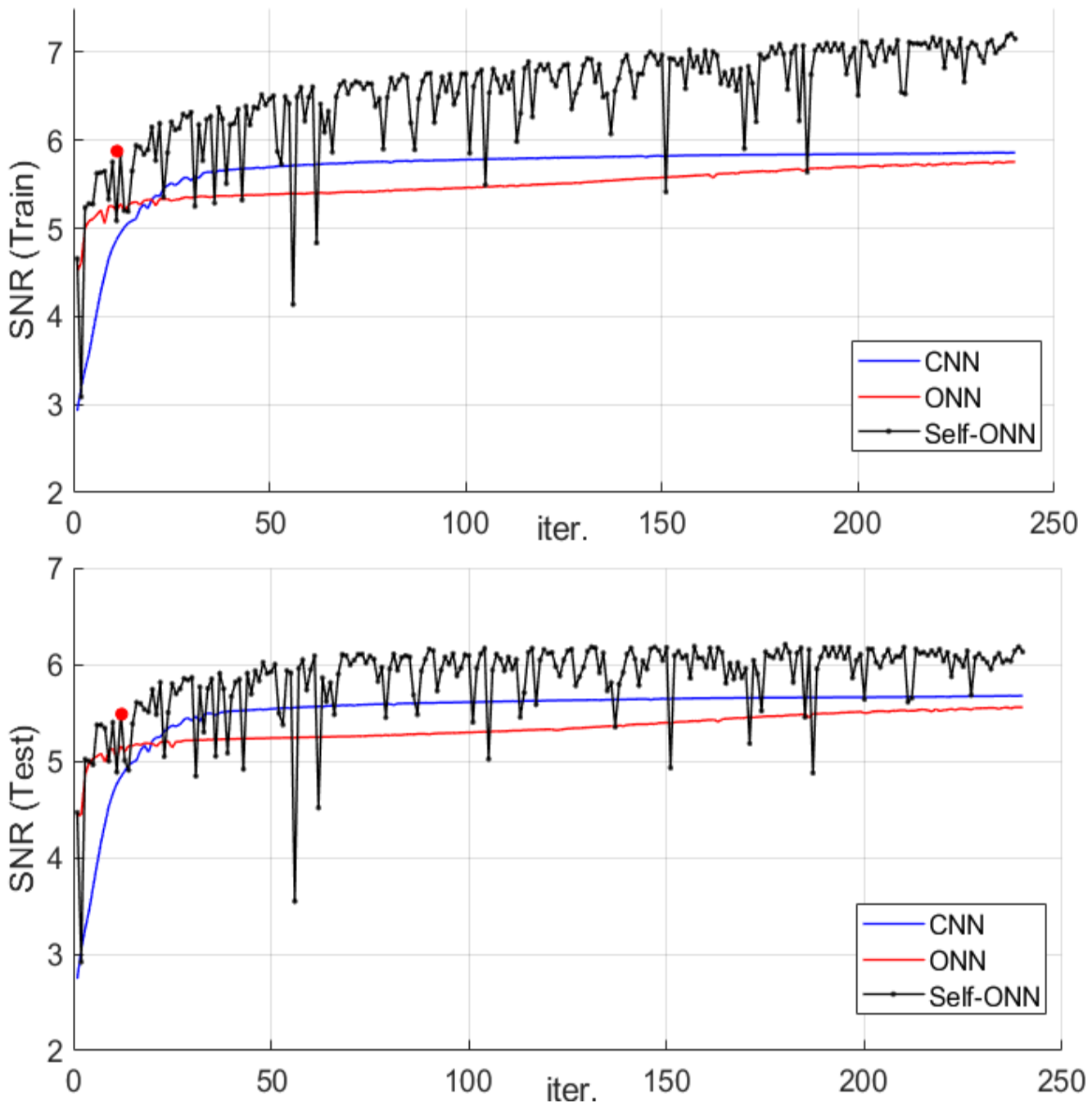

**Figure 5: The SNR vs. iteration plots for the CNN (blue), the ONN (red) and the Self-ONN (black) trained in the 1st fold. The red circle shows the maximum SNR level achieved by the competing networks.**

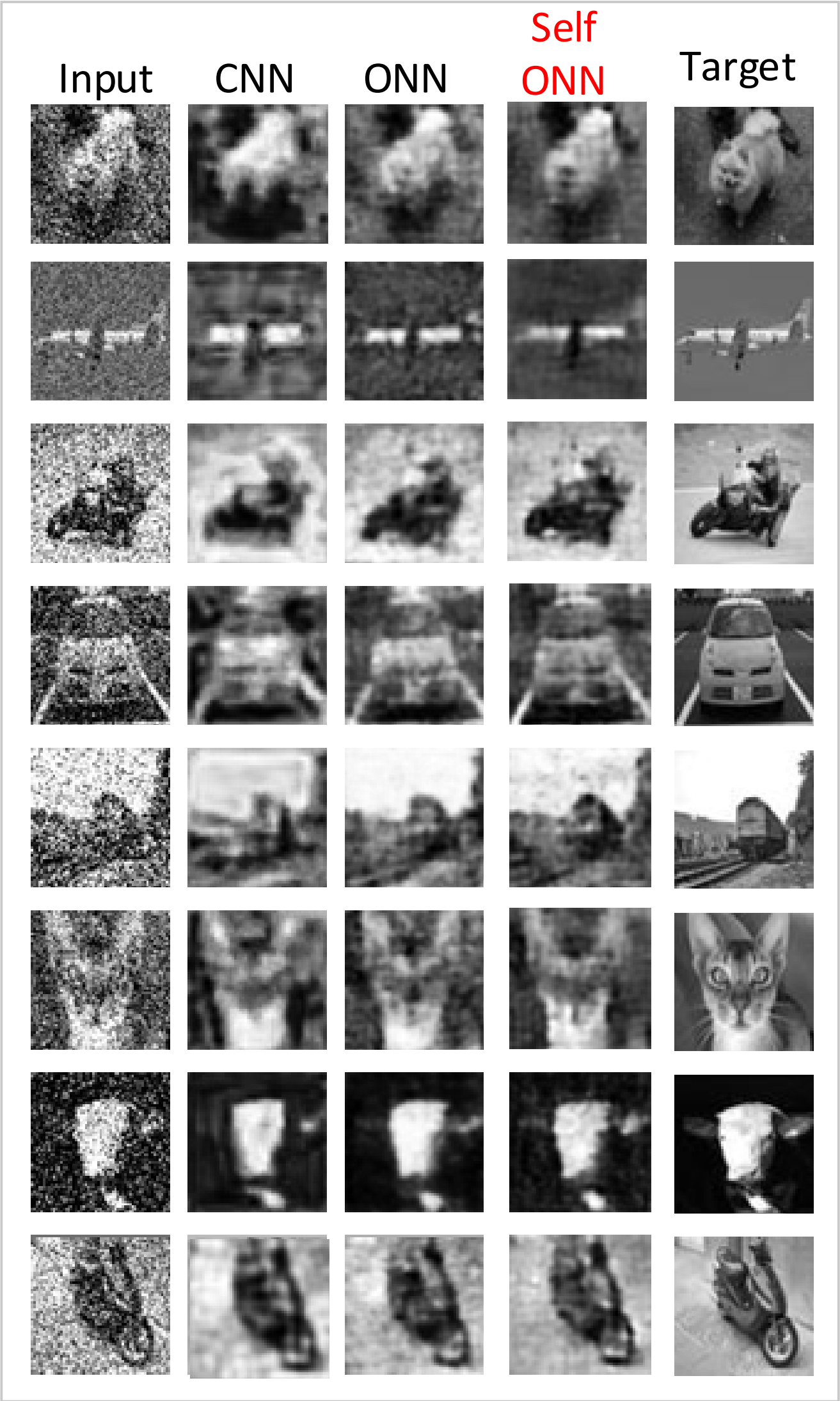

**Figure 6: Some random original (target) and noisy (images) and the corresponding outputs of the CNN, ONN and Self-ONN from the test partition.**

For a visual evaluation, Figure 6 shows randomly selected original (target) and noisy (input) images and the corresponding outputs of CNNs, ONNs and Self-ONNs from the test partition. The superior denoising performance of Self-ONNs is clear when compared with both traditional networks.

### 2) *Image Synthesis*

Image synthesis is a typical regression problem where a single network learns to synthesize a set of images from individual noise (WGN) images. As recommended in [37] we have trained a Self-ONN to (learn to) synthesize 8 (target) images from 8 WGN (input) images, as illustrated in Figure 7. We repeat the experiment 10 times (folds), so 8x10=80 images are randomly selected from Pascal VOC dataset. The gray-scaled and down-sampled original images are the target outputs while the WGN images are the input.

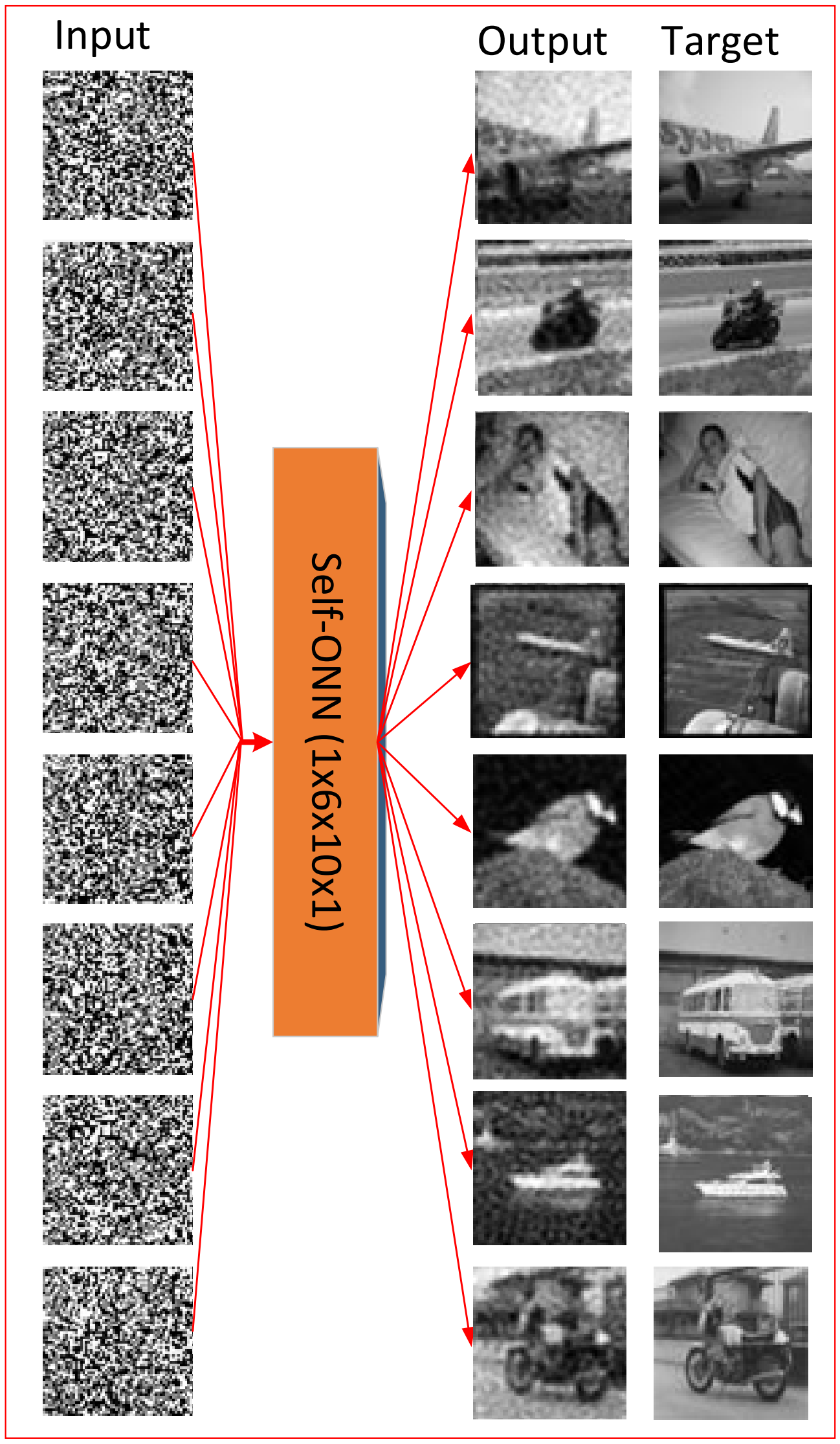

**Figure 7: The outputs of the BP-trained ONN with the corresponding input (WGN) and target (original) images from the 2nd synthesis fold.**

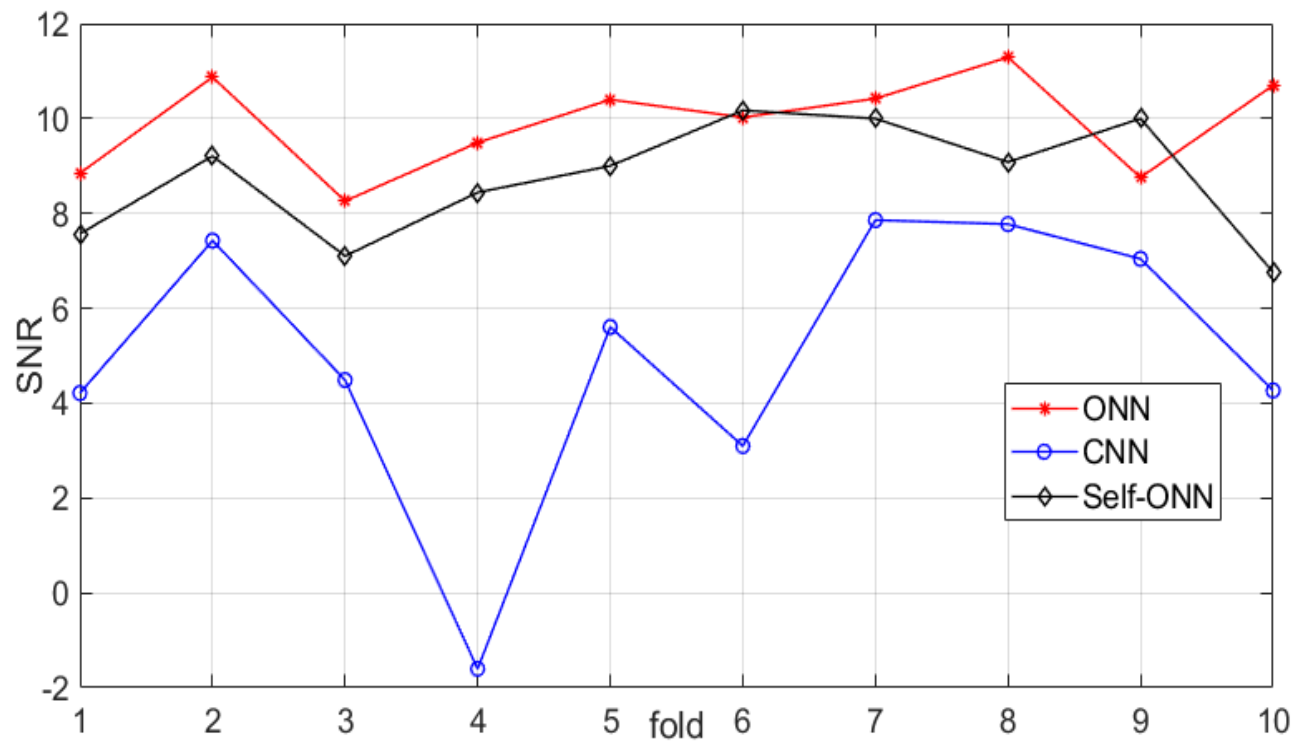

**Figure 8: Best SNR levels for each synthesis fold achieved by Self-ONN (black), CNNs (blue) and ONNs (red).**

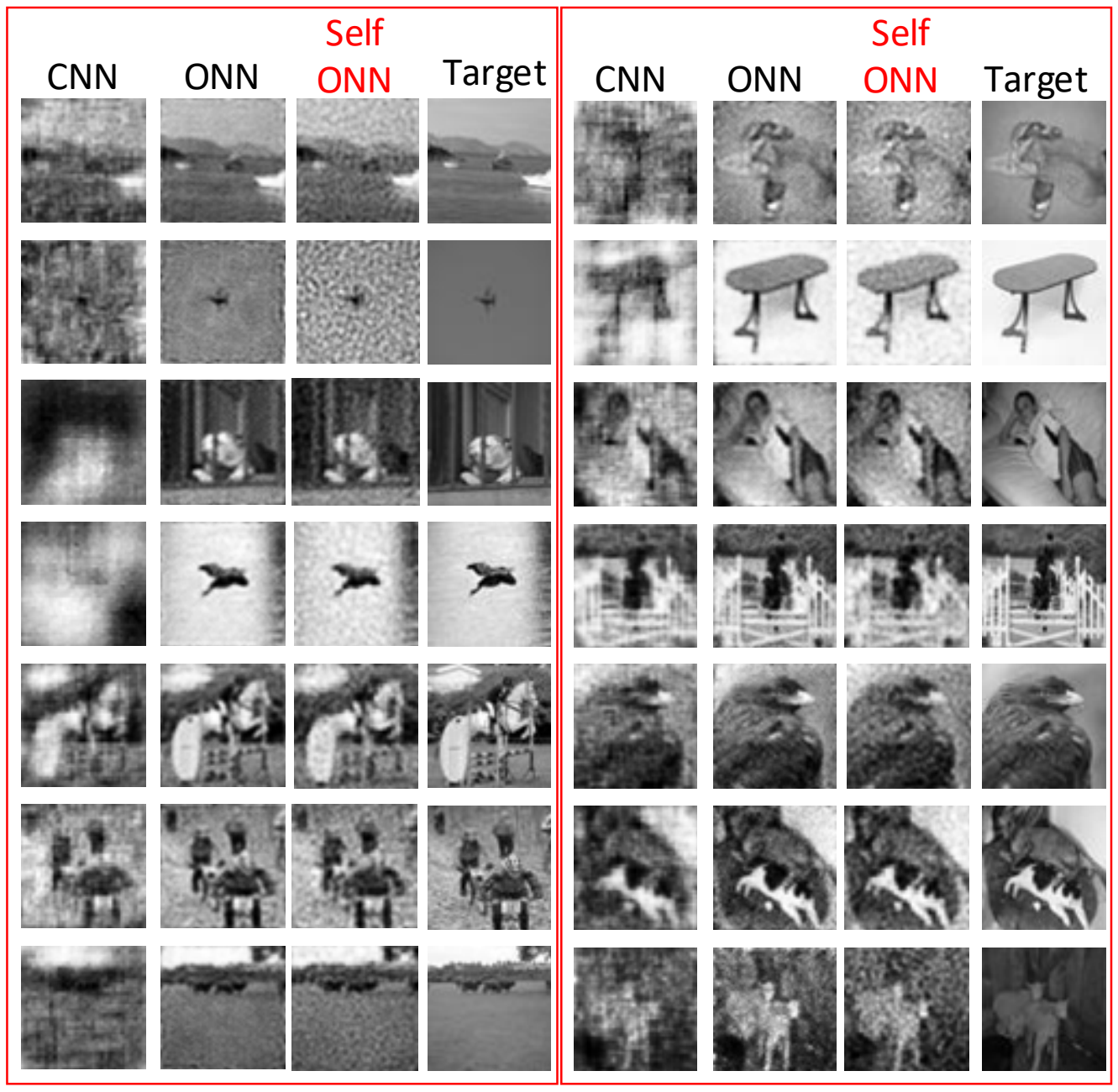

**Figure 9: A random set of 14 synthesis outputs of the best networks with the target images. The WGN input images are omitted.**

Figure 8 shows the SNR plots of CNNs, ONNs and Self-ONNs among the 10 BP runs for each synthesis experiment (fold). In this problem, Self-ONNs surpassed ONNs only on two folds out of ten. The average SNR levels of CNNs, ONNs and Self-ONNs synthesis are 5.02dB, 9.91dB, and 8.73dB respectively. The superiority of ONNs over Self-ONNs is due to two reasons: 1) The conventional nodal operators (exponential and chirp for the $1^{st}$ and $2^{nd}$ hidden layers and convolution for the output layer) are near-optimal choices whereas their Maclaurin approximation in Self-ONNs is not in general improving, rather deteriorating the learning performance, 2) conventional ONNs have the advantage of having 3 times more learning units (neurons) than Self-ONNs. Under the equivalent configuration, *1x16x32x1*, Self-ONNs still surpasses ONNs achieving an average SNR level of 10.27dB. Against CNNs, Self-ONNs demonstrate a superior performance with a significant average SNR gap over 3.5dB. Finally, for a visual comparative evaluation, Figure 9 shows a random set of 14 synthesis outputs of all networks with the target image. The performance gap is also clear here especially some of the CNN synthesis outputs have suffered from severe blurring and/or textural artefacts.

### *3) Face Segmentation*

Deep CNNs have often been used in face and object segmentation tasks [43]-[52]. As in [37], we used the benchmark FDDB face detection dataset [53], which contains 1000 images with one or many human faces in each image.

Figure 10 shows F1 plots of the best CNNs, ONNs and Self-ONNs at each fold over both partitions. ONN-3 is the ONN model that got the highest test F1 scores in [37]. The average F1 scores of CNN, ONN-3 and Self-ONN segmentation for the (train) and (test) partitions are: (58.58%, 79.86%, and 96.6%) and (56.74%, 59.61% and 62%), respectively. The first and the foremost interesting observation is that Self-ONNs can achieve significantly higher F1 level in train set despite the fact that both ONNs and CNNs have three times more learning units than Self-ONNs. In fact, such a train performance hints a certain amount of "over-fitting" which will be discussed next. Self-ONNs achieves the highest average F1 score on the test set too; however, the performance gap diminishes.

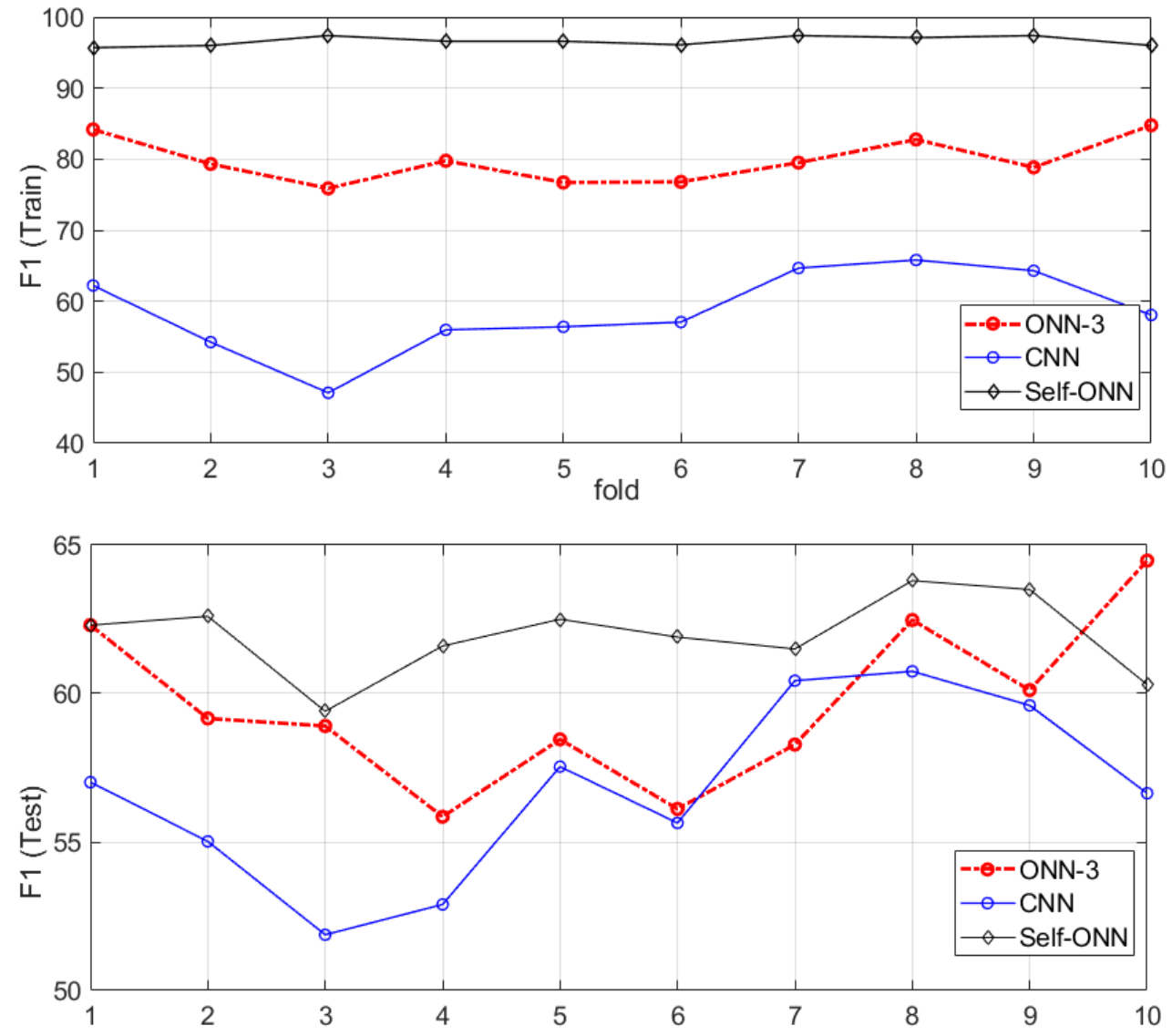

**Figure 10: Best segmentation F1 levels for each fold achieved in train (top) and test (bottom) partitions by Self-ONN (solid-black), ONN-3 (dashed-red) and CNN (solid-blue).**

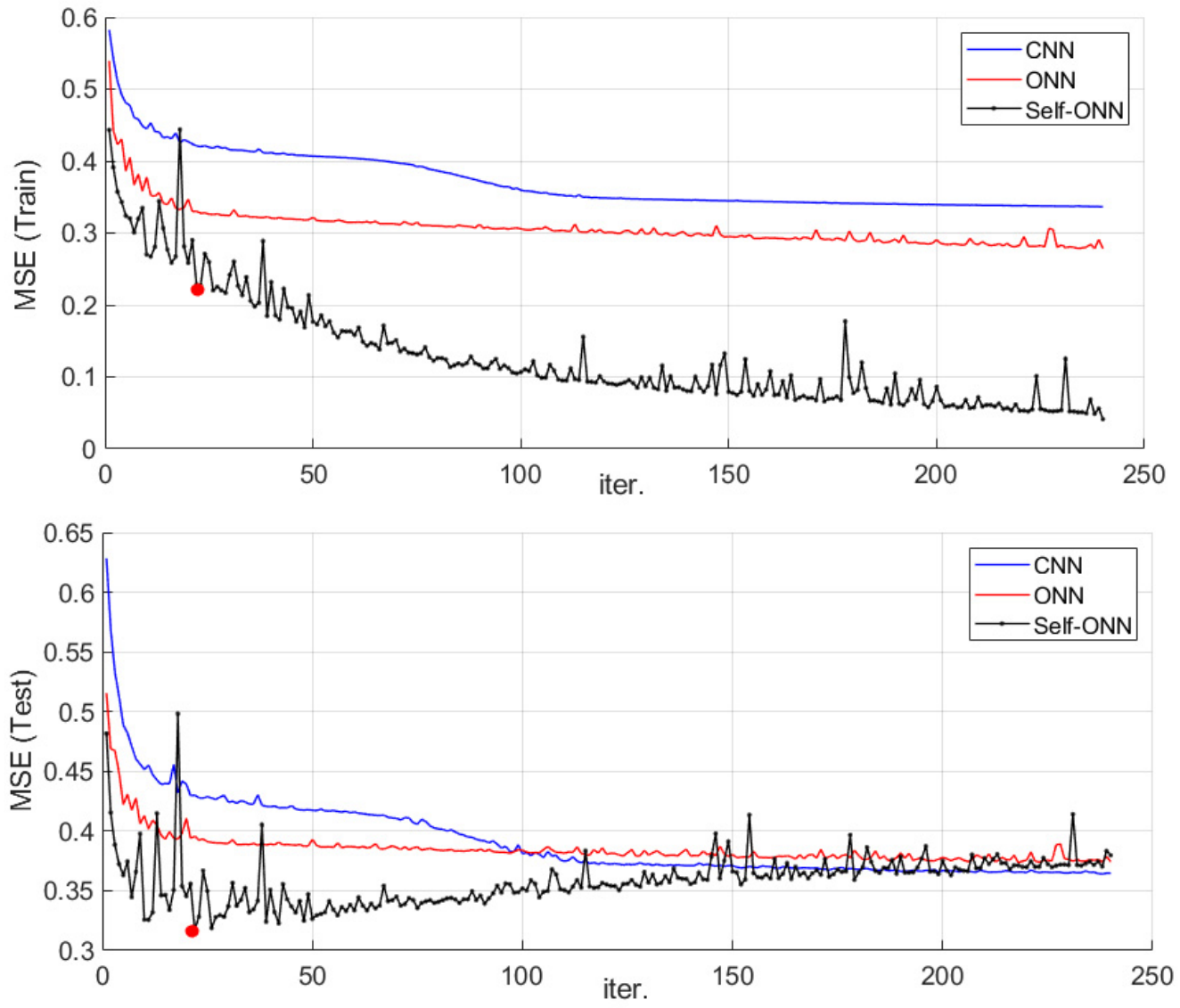

**Figure 11: The network loss (MSE) vs. iteration plots for the CNN (blue), the ONN (red) and the Self-ONN (black) trained in the $1^{st}$ fold. The red circle shows the minimum loss achieved on the test set.**

Figure 11 presents the loss (MSE) vs. iteration curves of all networks for the $1^{st}$ fold. As in the denoising problem, the Self-ONN shows a staggering convergence speed, i.e., on both train and test sets, Self-ONN can achieve the minimum loss level of both CNN and ONN only within 10 iterations whilst both competing networks achieve their minimum loss almost at the end of the training. Then it reaches the minimum loss (MSE = 0.324) at iteration 21 and thereafter, the loss gradually increases at the test set, which indicates an over-fitting. This is not surprising considering the scarcity of the train data and the lesser number of learning units in SelfONNs. In practice, such an overfitting can be avoided with a standard "early-stopping" technique over a validation set, and this, in turn, allows a very brief BP training (e.g. < 50 iterations) to achieve an elegant learning performance on the test set.

*4) Image Transformation*

In this task, a set of images is transformed to another by a network. In all earlier image transformation applications of Deep CNNs [54], [55] the input and output images are strongly correlated, e.g., edge-to-image, gray-scale-to-color image, and day-to-night (or vice versa) photo translation, etc. In [37] this problem has become more challenging where each image is transformed to an entirely different image. Moreover, a single network is trained to (learn to) transform 4 (target) images from 4 input images, as illustrated in Figure 12 (left). In this fold, note that two pairs of distinct images are used as both input and output of each other; therefore, the capability of the networks to learn both "forward" and "backward" problems at the same time and for two image pairs is tested.

For this task, some images selected from the FDDB face detection dataset [53] are used. For comparison, we used the results obtained from CNNx4 configuration (*1x32x64x1*) since it achieved the best results reported in [37]. For comparison, the best ONN model recommended in [37] is used with the operator indices, 0 and 13 for the 1st and 2nd hidden layers corresponding to the operator indices: 0:{0, 0, 0} for the pool (*summation*=0), activation (*tanh*=0) and nodal (*mul*=0), respectively, and 13:{0,1,6} for the pool (*summation*=0), activation (*lin-cut*=1) and nodal (*chirp*=6), respectively.

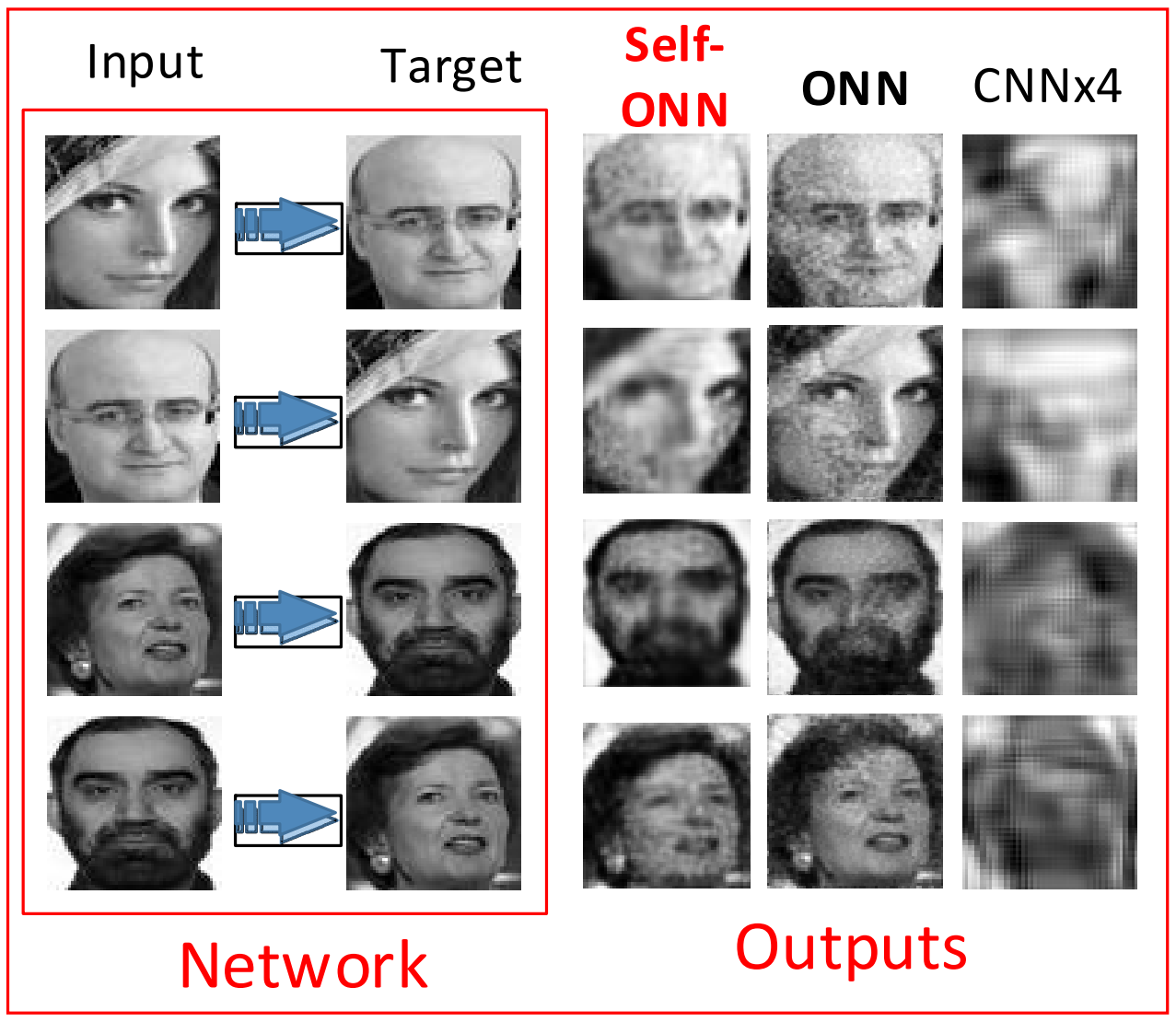

**Figure 12: Image transformation of the 1st fold including two inverse problems (left) and the outputs of the ONN and CNN with the default configuration, and the two CNNs (CNNx4) with 4 times more parameters. On the bottom, the numbers of input → target images are shown.**

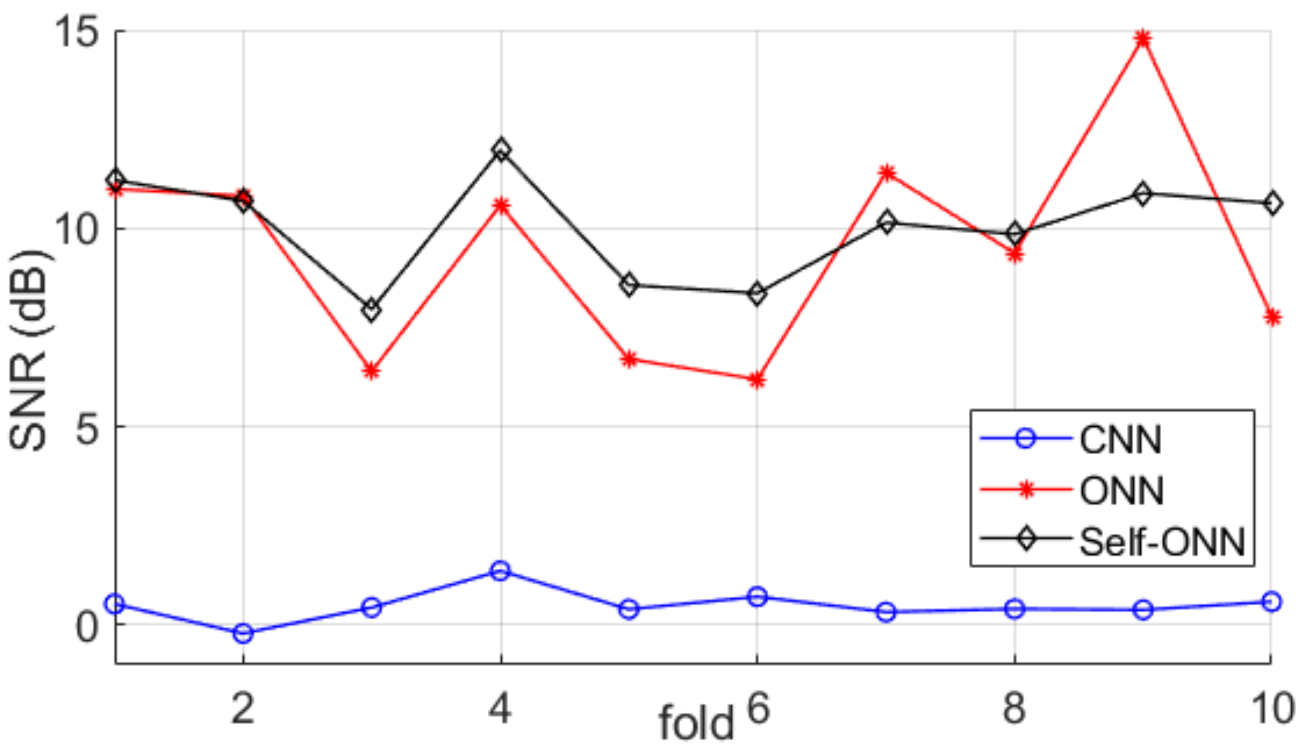

**Figure 13: Best SNR levels for each image transformation fold achieved by the corresponding Self-ONN (black), CNNx4 (blue) and ONN (red).**

Figure 13 presents the best SNR levels for each image transformation fold for all networks. The average SNR levels achieved by CNNs, ONNs and Self-ONNs are 0.5dB, 9.5dB and 10dB, respectively. As this is the hardest learning problem among all the problems in this study, it is not surprising to observe the largest performance gap between CNN and both ONN models (higher than 9dB on the average). The performances of ONNs and Self-ONNs are within a narrow margin while ONNs surpassed Self-ONNs on 3 out of 10 transformations. A close look to the plots in Figure 13 reveals the fact that in fold 9, the ONN has significantly surpassed the Self-ONN (as in fold 10 in image synthesis problem). This happens when the nodal operator of each neuron fits very well, and thus its $Q^{th}$ order approximation cannot reach the same performance. Moreover, the Maclaurin approximation also costs $Q$-times more parameters and thus, the Self-ONN ends up with significantly less number of neurons, which can potentially deteriorate the learning performance. However, more often Self-ONNs can surpass ONNs when their nodal operators are not properly assigned, or more likely, no such "best-fitting" nodal operator is available in the operator set library for the problem at hand. Obviously, in this case the "custom-made" nodal operators by generative neurons can boost the learning performance that is visible in the majority of the experiments performed in this study.

## B. *Computational Complexity Analysis*

In this section the computational complexity of the proposed Self-ONNs is analyzed with respect to the parameter-equivalent CNNs and ONNs. We shall begin with the complexity analysis of the forward propagation (FP) and then focus on BP. For the sake of simplicity, we shall ignore the up- and down-sampling and assume the same input map sizes among the layers.

As assumed in this study, when the pool operator is "sum", $P_i^l = \Sigma$, in a FP in a Self-ONN, Eq. (1) can be expressed as follows:

$$x_k^l = b_k^l + \sum_{i=1}^{N_{l-1}} oper2D(w_{ki}^l, y_i^{l-1}, 'NoZeroPad')$$
$$x_k^l(m,n)\Big|_{(0,0)}^{(M-1,N-1)} = b_k^l + \sum_{i=1}^{N_{l-1}} \left( \sum_{r=0}^{K_x-1} \sum_{t=0}^{K_y-1} \boldsymbol{\Psi}\left(\boldsymbol{w}_{ki}^l(\mathbf{r},\mathbf{t}), y_i^{l-1}(m+r, n+t)\right) \right) \quad (20)$$

where $\boldsymbol{\Psi}$ is the (Maclaurin) composite nodal operator and $\boldsymbol{w}_{ki}^{l}(\mathbf{r}, \mathbf{t})$ is Q-dimensional array for the kernel element $(\mathbf{r}, \mathbf{t})$. Putting the $q^{\text{th}}$ order 2D kernel, $w_{ki}^{l}\langle q\rangle$ ($q$=1..$Q$), which is composed of the *kernel elements,* $w_{ik}^{l+1}(\mathrm{r}, \mathrm{t}, q)$, then Eq. (20) can be simplified as,

$$x_k^l = b_k^l + \sum_{q=1}^{Q}\left\{\sum_{i=1}^{N_{l-1}} conv2D\left(w_{ki}^{l}\langle q\rangle, \left(y_i^{l-1}\right)^q, 'NoZeroPad'\right)\right\} \quad (21)$$

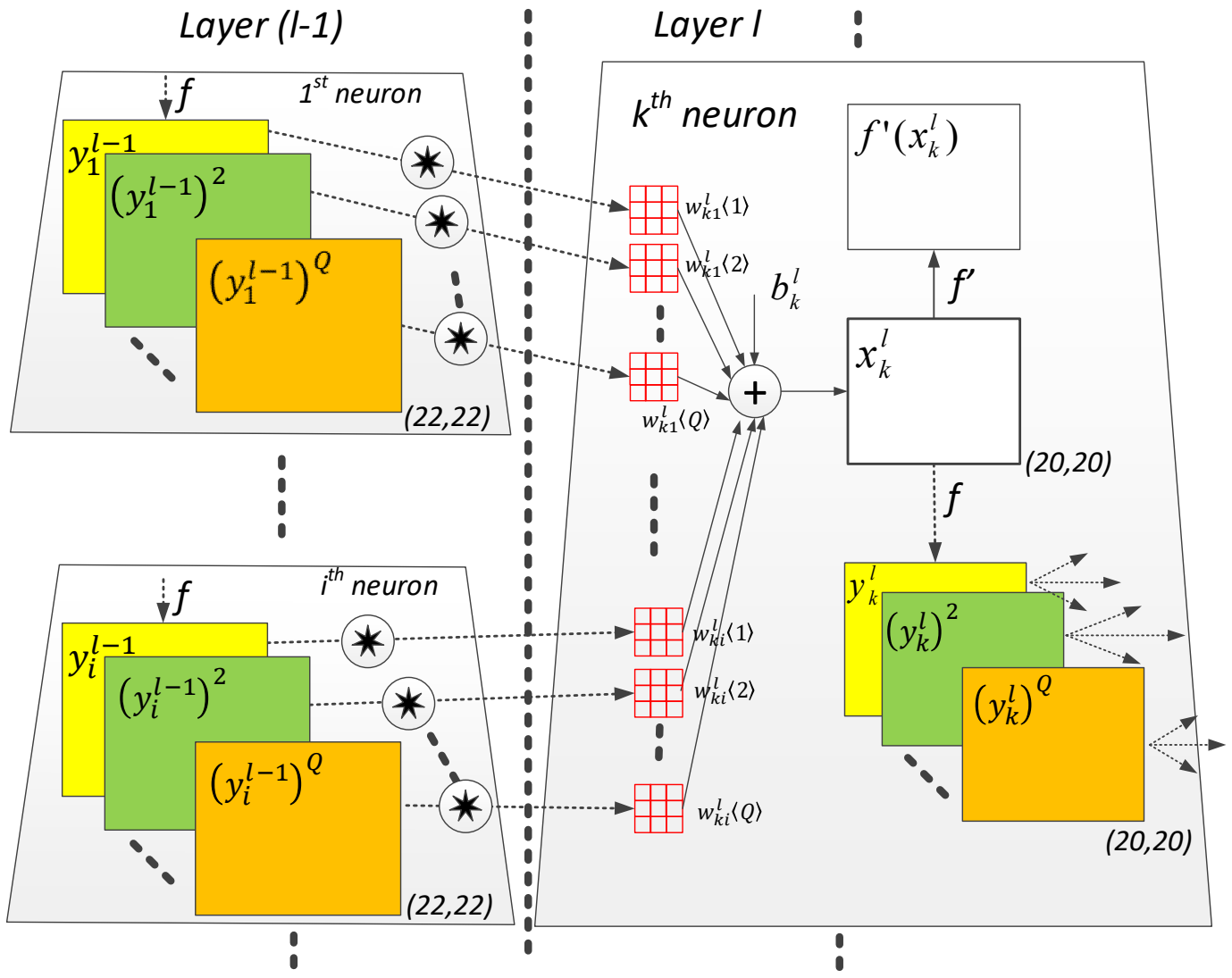

**Figure 14: The illustration of a Self-ONN equivalent to Figure 2 (right) when the pool operator is "sum", $\mathbf{P}_i^l = \boldsymbol{\Sigma}$, and the activation function is *tanh*.**

This special-case Self-ONN configuration is illustrated in Figure 14 where it actually resembles a multi-output and multi-kernel CNN. Once the power-outputs, $\left(y_i^{l-1}\right)^q$, for $q$=1..$Q$, are computed for all hidden neurons in the network, Eq. (21) is simply $(Q \times N_{l-1})$ independent 2D convolutions, which can be parallelized, and hence, will take the same time for a single convolution. Therefore, in a parallelized implementation, a Self-ONN and a CNN with the same configuration have approximately the same computational complexity for FP. In this study, we compute the total number of multiply-accumulate operations (MACs) for CNNs and Self-ONNs used in this study. The number of MACs for the $l^{th}$ layer of the network is calculated using the following formula:

$$MACs(l) = |Y_l| * \left(\left(N_{l-1} * K_x^l * K_y^l * Q^l\right) + N_l\right)$$

where $Y_l$ is the output of the current layer, $N_{l-1}$ is the number of neurons in the previous layer, $K_x^l$ and $K_y^l$ are the kernel dimensions for the current layer, $Q^l$ is the order of approximation and finally,$N_l$ is the number of neurons in the current layer. The last term can be omitted for the special case where bias is not used. Table 1 provides the comparisons of the number of trainable parameters and total number of MACs, of the networks used in this study. The number of neurons in the input and output is fixed to 1 for both networks.

**Table 1 Comparison of total number of multiply accumulate operations for the networks used in this study.**

| Network | Layer 1 Neurons | Layer 2 Neurons | Trainable Parameters (k) | Total MACs (M) |
|---|---|---|---|---|
| **SelfONN** | 6 | 10 | 39.749 | 78.246 |
| **CNN** | 16 | 32 | 32.481 | 79.923 |

For computational complexity comparison in BP training, recall that when the pool operator is "sum", then $\nabla_{\Psi_{ki}} \mathrm{P}_i^{l+1} = 1$. Recall further that the power outputs, $\left(y_i^{l-1}\right)^q$, are already computed for each hidden neuron of the network during the prior FP for each BP iteration. So, once the 4D matrix, $\nabla_y \boldsymbol{\Psi}$, is computed by Eq. (11), then the error back-propagation computation in Eq. (10) can be parallelized and will take the same time complexity with an equivalent CNN. The computation of $f'\left(x_k^l\right)$, and the delta errors, $\Delta_k^l$, are also common with the conventional BP in CNNs. This makes the identical computational complexity for bias sensitivities by Eq. (19). Finally, for weight sensitivities, note that Eq. (18) is simply Q independent convolutions of the delta error, $\Delta_i^{l+1}$, and power output, $\left(\mathrm{y}_\mathrm{k}^\mathrm{l}\right)^\mathrm{q}$, all of which can also be parallelized to take the same time for a single convolution. As a result, there is no significant difference between the BP computational complexities of CNNs and Self-ONNs with the same configuration. The time for computing the power outputs (only once in each BP iteration) and the 4D matrices are the only overheads, which are insignificant. In this study, earlier analogy is also valid for BP, i.e., since the network configuration for Self-ONN has three times less neurons than ONNs and CNNs, BP for Self-ONNs will take significantly less time than the BP for CNNs. The gap further widens when compared to ONNs about 1.5 to 4.7 times in practice [37].

## V. Conclusions

In this study, Self-Organized ONNs (Self-ONNs) are proposed with the generative neuron model, which allows customized (self-optimized) nodal operator functions -not only for each neuron but for each kernel connection to the previous layer neurons. This is an ultimate heterogeneity level that allows to create (self-) optimized nodal operators during BP training. This does not only void the need for prior operator search runs, but also optimizes the nodal operator of the output layer neuron(s), that are the most crucial neurons in the network in which the loss (fitness) is computed. Like ONNs, Self-ONNs are also a superset of CNNs, e.g., when the order of a Self-ONN for each layer is set to $Q$=1, a Self-ONN will become a CNN. Even when $Q > 1$, if a linear (convolutional) neuron is the optimal choice for a particular problem, the ongoing BP can still converge all higher order ($q$>1) weights to zero to turn the Self-ONN to a conventional CNN. Overall, the generative neurons have the ability to form customized nodal operators per kernel connection for the problem at hand. In this way, traditional "weight optimization" of conventional CNNs is now turned to be "operator optimization" process.

The results on the four challenging problems proposed in [37] show that Self-ONNs with the same number of parameters (but with much less number of neurons) can achieve a superior learning performance whilst the performance gap over CNNs widens further. Self-ONNs usually obtain comparable or better results than ONNs; however, some results have highlighted a crucial fact: when a conventional nodal operator of an ONN is the "right choice" for a particular problem, the parameter-equivalent Self-ONN cannot surpass the performance with the $Q$-order Maclaurin approximation of the "near-optimal" nodal operators and of course, with less number of neurons. However, this seems to be the minority case over the problems tackled in this study. Above all, Self-ONNs, in a parallelized implementation, have a superior computational efficiency especially compared to ONNs.

This study has proposed a "baseline" version of Self-ONNs and further performance boost can be expected with the following improvements:

- instead of fixing to some practical value, (e.g. $Q$=7 in this study) optimizing the order of Maclaurin approximation, $Q$, per layer and even per neuron,
- adapting a better optimization scheme for training, e.g., SGD with momentum [56], AdaGrad [57], RMSProp [58], Adam [59] and its variants [60], all of which should be modified for Self-ONNs for proper functioning,
- optimizing also the pool and activation operators during BP training,
- and finally, performing non-localized kernel operations for each kernel connection of each neuron for the operation capability within a larger area *without* increasing the size of the kernels.

These will be the topics for our future research.

## APPENDIX

### A. BP Training for Operational Neural Networks

As mentioned earlier, conventional Back-Propagation (BP) training consists of four consecutive stages: 1) Computation of the delta error, $\Delta_1^L$, at the output layer, 2) Inter-BP between two consecutive operational layers, 3) Intra-BP in an operational neuron, and 4) Computation of the weight (operator kernel) and bias sensitivities in order to update them at each BP iteration. The following sub-sections will explain each stage in detail. For the sake of brevity, a Stochastic Gradient Descent-based minimization of the L2-loss is assumed. In practice, any first order differentiable error function can be used.

#### 1) Computation of the delta error, $\Delta_1^L$, at the output layer

Typically, the L2-loss or the Mean-Square-Error (MSE) error function for an image $I$ in the train dataset can be expressed as follows .

$$E(I) = \sum_p \left(y_1^L(I_p) - T(I_p)\right)^2 \tag{22}$$

where $I_p$ is the pixel $p$ of the image $I$, $T$ is the target output and $y_1^L$ is the predicted output. The delta sensitivity of the error can then be computed as:

$$\Delta_1^L = \frac{\partial E}{\partial x_1^L} = \frac{\partial E}{\partial y_1^L}\frac{\partial y_1^L}{\partial x_1^L} = \frac{2}{|I|}\left(y_1^L(I) - T(I)\right) f'\left(x_1^L(I)\right) \tag{23}$$

#### 2) Inter-BP between two operational layers: $\Delta y_k^l \overset{\Sigma}{\leftarrow} \Delta_i^{l+1}$

We first focus on the contribution of a single output pixel, $y_k^l(m,n)$, to the pixels of the $x_i^{l+1}$. Assuming again a $K_x \times K_y = 3 \times 3$ kernel, Eq. (24) formulates the contribution of $y_k^l(m,n)$ to the 9 neighboring pixels. According to the basic rule of BP one can then formulate the delta of $y_k^l(m,n)$ as in Eq. (25). Note that the output pixel, $y_k^l(m,n)$, and input pixel, $x_i^{l+1}(m,n)$, are connected through the first (top-left) element of the kernel, $x_i^{l+1}(m,n) = \ldots + P_i^{l+1}\left[\Psi_i^{l+1}\left(y_k^l(m,n), w_{ik}^{l+1}(0,0)\right), \ldots, \Psi_i^{l+1}\left(y_k^l(m+r, n+t), \mathrm{w}_{\mathrm{ik}}^{\mathrm{l}}(\mathrm{r},\mathrm{t}), \right)\ldots\right)\right]$. This means that the contribution of $y_k^l(m,n)$, will now only be on $x_i^{l+1}(m-r, n-t)$ as expressed explicitly in Eq. (24). The chain-rule of derivatives should now include the two operator functions, pool and nodal, which, in case of CNNs, are fixed to summation and multiplication respectively. The delta error of the output pixel can, therefore, be expressed as in Eq. (25) in the generic form of pool, $P_i^{l+1}$, and nodal, $\Psi_i^{l+1}$, operator functions of each operational neuron $i \in [1, .., N_{l+1}]$ in the next layer.

$$\begin{aligned}
x_i^{l+1}(m-1,n-1) &= \ldots + P_i^{l+1}[\Psi_i^{l+1}(y_k^l(m-1,n-1), w_{ik}^{l+1}(0,0)), \ldots, \Psi_i^{l+1}(y_k^l(m,n), w_{ik}^{l+1}(1,1))] + \ldots \\
x_i^{l+1}(m-1,n) &= \ldots + P_i^{l+1}[\Psi_i^{l+1}(y_k^l(m-1,n), w_{ik}^{l+1}(0,0)), \ldots, \Psi_i^{l+1}(y_k^l(m,n), w_{ik}^{l+1}(1,0)), \ldots] + \ldots \\
x_i^{l+1}(m,n) &= \ldots + P_i^{l+1}[\Psi_i^{l+1}(y_k^l(m,n), w_{ik}^{l+1}(0,0)), .., \Psi_i^{l+1}(y_k^l(m+r,n+t), w_{ik}^{l+1}(\mathrm{r},\mathrm{t}), ) \ldots)] + \ldots \\
&\ldots\ldots \\
x_i^{l+1}(m+1,n+1) &= \ldots + P_i^{l+1}[\Psi_i^{l+1}(y_k^l(m+1,n+1), w_{ik}^{l+1}(0,0)), \ldots] + \ldots \\
\therefore\ x_i^{l+1}(m-r,n-t)\Big|_{(1,1)}^{(M-1,N-1)} &= b_i^{l+1} + \sum_{k=1}^{N_1} P_i^{l+1}[\ldots, \Psi_i^{l+1}(w_{ik}^{l+1}(r,t), y_k^l(m,n)), \ldots]
\end{aligned} \tag{24}$$

$$\therefore\ \frac{\partial E}{\partial y_k^l}(m,n)\Bigg|_{(0,0)}^{(M-1,N-1)} = \Delta y_k^l(m,n) =$$
$$\sum_{i=1}^{N_{l+1}}\left(\sum_{r=0}^{\mathrm{Kx}-1}\sum_{t=0}^{\mathrm{Ky}-1} \frac{\partial E}{\partial x_i^{l+1}(m-r,n-t)} \times \frac{\partial x_i^{l+1}(m-r,n-t)}{\partial P_i^{l+1}[.., \Psi_i^{l+1}(y_k^l(m,n), w_{ik}^{l+1}(r,t)), ..]} \times \frac{\partial P_i^{l+1}[.., \Psi_i^{l+1}(y_k^l(m,n), w_{ik}^{l+1}(r,t)), ..]}{\partial \Psi_i^{l+1}(y_k^l(m,n), w_{ik}^{l+1}(r,t))} \times \frac{\partial \Psi_i^{l+1}(y_k^l(m,n), w_{ik}^{l+1}(\mathrm{r},\mathrm{t}))}{\partial y_k^l(m,n)}\right) \tag{25}$$

In Eq. (25), note that the first term, $\frac{\partial x_i^{l+1}(m-r,n-t)}{\partial P_i^{l+1}[..,\Psi_i^{l+1}(y_k^l(m,n),w_{ik}^{l+1}(r,t)),..]} = 1$. Let $\nabla_{\Psi_{ki}} \mathrm{P}_i^{l+1}(m,n,r,t) = \frac{\partial P_i^{l+1}[..,\Psi_i^{l+1}(y_k^l(m,n),w_{ik}^{l+1}(r,t)),..]}{\partial \Psi_i^{l+1}(y_k^l(m,n),w_{ik}^{l+1}(r,t))}$ and $\nabla_y \Psi_{ki}^{l+1}(m,n,r,t) = \frac{\partial \Psi_i^{l+1}(y_k^l(m,n),w_{ik}^{l+1}(r,t))}{\partial y_k^l(m,n)}$. First, it is obvious that both derivatives, $\nabla_{\Psi_{ki}} \mathrm{P}_i^{l+1}$, and $\nabla_y \Psi_{ki}^{l+1}$, no longer require the rotation of the kernel, $w_{ik}^{l+1}$. The first derivative, $\nabla_{\Psi_{ki}} \mathrm{P}_i^{l+1}$, depends on the role (contribution) of the particular nodal term, $\Psi_i^{l+1}(y_k^l(m,n), w_{ik}^{l+1}(r,t))$, within the pool function. The derivative, $\nabla_{\Psi_{ki}} \mathrm{P}_i^{l+1}(m,n,r,t)$ is computed while computing the pixels $x_i^{l+1}(m-r,n-t)$ for $\forall r,t \in (Kx, Ky)$ that corresponds to the particular output value, $y_k^l(m,n)$, within each pool function. Recall that this is the contribution of the $y_k^l(m,n)$ alone for each input value at the next layer, $x_i^{l+1}(m-r,n-t)$ for $\forall r,t \in (Kx, Ky)$. When the pool operator is summation, $\mathrm{P}_i^{l+1} = \Sigma$, then $\nabla_{\Psi_{ki}} P_i^{l+1} = 1$, which is constant for any nodal term. For any other alternative, the derivative $\nabla_{\Psi_{ki}} \mathrm{P}_i^{l+1}(m,n,r,t)$ will be a function of four variables. The other derivative term, $\nabla_y \Psi_{ki}^{l+1}$, is the derivative of the nodal operator with respect to the output. For the simplest case of a linear neuron which has the "multiplication" nodal operator, i.e., $\Psi_i^{l+1}(y_k^l(m,n), w_{ik}^{l+1}(r,t)) = y_k^l(m,n).w_{ik}^{l+1}(r,t)$, the derivative $\nabla_y \Psi_{ki}^{l+1}$ is simply the weight kernel, $w_{ik}^{l+1}(r,t)$. and is independent of the output, $y_k^l(m,n)$. But as a general rule, the derivative $\nabla_y \Psi_{ki}^{l+1}(m,n,r,t)$ will also be a function of four variables. By using these four variable derivatives or equivalently, the two 4-D matrices, Eq. (25) can be simplified as Eq. (26). Note that $\Delta y_k^l$, $\nabla_{\Psi_{ki}} \mathrm{P}_i^{l+1}$ and $\nabla_y \Psi_{ki}^{l+1}$ have the size, $M \times N$ while the next layer delta error, $\Delta_\mathrm{i}^{l+1}$, has the size, $(M-\mathrm{K_x}+1)\times(N-\mathrm{K_y}+1)$, respectively. Therefore, to make the variable 2D convolution in this equation valid, the delta error, $\Delta_\mathrm{i}^{l+1}$, is symmetrically padded with $\mathrm{K_x}-1$ and $\mathrm{K_y}-1$ zeros across the width and height respectively.

$$\Delta y_k^l(m,n)\Big|_{(0,0)}^{(M-1,N-1)}=\sum_{i=1}^{N_{l+1}}\left(\sum_{r=0}^{\mathrm{K_x}-1}\sum_{t=0}^{\mathrm{K_y}-1}\Delta_\mathrm{i}^{l+1}(m-r,n-t)\times\nabla_{\Psi_{ki}}\mathrm{P}_i^{l+1}(m,n,r,t)\times\nabla_y\Psi_{ki}^{l+1}(m,n,r,t)\right)$$
$$Let\ \nabla_y\mathrm{P}_i^{l+1}(m,n,r,t)=\nabla_{\Psi_{ki}}\mathrm{P}_i^{l+1}(m,n,r,t)\ \times\nabla_y\Psi_{ki}^{l+1}(m,n,r,t), then \tag{26}$$
$$\Delta y_k^l=\sum_{i=1}^{N_{l+1}}Conv2Dvar\left\{\Delta_\mathrm{i}^{l+1},\nabla_y\mathrm{P}_i^{l+1}(m,n,r,t)\right\}$$

*3) Intra-BP in an operational neuron:* $\Delta_k^l\overset{BP}{\leftarrow}\Delta y_k^l$

If there is no subsampling performed within the neuron, once the delta-errors are back-propagated from the next layer, *l*+1, to the neuron in layer, *l*, then we can further back-propagate it to the input delta. This can be formulated as follows:

$$\Delta_k^l=\frac{\partial E}{\partial x_k^l}=\frac{\partial E}{\partial y_k^l}\frac{\partial y_k^l}{\partial x_k^l}=\frac{\partial E}{\partial y_k^l}f'(x_k^l)=\Delta y_k^l\,f'(x_k^l) \tag{27}$$

where $\Delta y_k^l$ is computed as in Eq. (26). On the other hand, if there is a down-sampling by factors, *ssx* and *ssy*, then the back-propagated delta-error by Eq. (26) should be first up-sampled to compute the delta-error of the neuron. Let zero order up-sampled map be: $uy_k^l=\underset{ssx,ssy}{\mathrm{up}}(y_k^l)$. Then Eq. (27) can be updated as follows:

$$\Delta_k^l=\frac{\partial E}{\partial x_k^l}=\frac{\partial E}{\partial y_k^l}\frac{\partial y_k^l}{\partial x_k^l}=\frac{\partial E}{\partial y_k^l}\frac{\partial y_k^l}{\partial uy_k^l}\frac{\partial uy_k^l}{\partial x_k^l}$$
$$=\underset{ssx,ssy}{\mathrm{up}}(\Delta y_k^l)\beta\,f'(x_k^l) \tag{28}$$

where $\beta=\frac{1}{ssx.ssy}$ since each pixel of $y_k^l$ is now obtained by averaging (*ssx.ssy*) number of pixels of the intermediate output, $uy_k^l$. Finally, if there is an up-sampling by factors, *usx* and *usy*, then let the average-pooled map be: $dy_k^l=\underset{usx,usy}{\mathrm{down}}(y_k^l)$. Then Eq. (27) can be updated as follows:

$$\Delta_k^l=\frac{\partial E}{\partial x_k^l}=\frac{\partial E}{\partial y_k^l}\frac{\partial y_k^l}{\partial x_k^l}=\frac{\partial E}{\partial y_k^l}\frac{\partial y_k^l}{\partial dy_k^l}\frac{\partial dy_k^l}{\partial x_k^l}$$
$$=\underset{usx,usy}{\mathrm{down}}(\Delta y_k^l)\beta^{-1}f'(x_k^l) \tag{29}$$

where $\beta=usx.usy$.

*4) Computation of the Weight (Kernel) and Bias Sensitivities*

The first three BP stages are performed to compute and back-propagate the delta errors, $\Delta_k^l=\frac{\partial E}{\partial x_k^l}$, to each operational neuron at each hidden layer. As illustrated in Figure 2, a delta error is a 2D map whose size is identical to the input map of the neuron. The sole purpose of back-propagating the delta-errors at each BP iteration is to use them to compute the weight and bias sensitivities.

First of all, the bias for the $k^{\text{th}}$ neuron in layer l, $b_k^l$, contributes to all pixels in the image (same bias shared among all pixels), so its sensitivity will be the accumulation of individual pixel sensitivities as expressed in Eq. (30):

$$\frac{\partial E}{\partial b_k^l}=\sum_{m=0}^{M-1}\sum_{n=0}^{N-1}\frac{\partial E}{\partial x_k^l(m,n)}\frac{\partial x_k^l(m,n)}{\partial b_k^l}$$
$$=\sum_{m=0}^{M-1}\sum_{n=0}^{N-1}\Delta_k^l\,(m,n) \tag{30}$$

Eq. (31) shows the contribution of bias and weights to the next level input map. $x_i^{l+1}(m,n)$. In order to derive the expression for the weight sensitivities we can follow the same approach as before: since each kernel element, $w_{ik}^{l+1}(r,t)$ affects *all* the pixels of the input map, $x_i^{l+1}$, by using the chain rule, the weight sensitivities can first be expressed as in Eq. (32) and then simplified into the final form in Eq. (33).

$$Recall\text{:}\ x_i^{l+1}(m-r,n-t)\Big|_{(\mathrm{Kx,Ky})}^{(M-1,N-1)}=b_i^{l+1}+\sum_{k=1}^{N_1}P_i^{l+1}[\ldots,\Psi_i^{l+1}(w_{ik}^{l+1}(r,t),y_k^l(m,n)),\ldots] \tag{31}$$

$$\therefore \frac{\partial E}{\partial w_{ik}^{l+1}}(r,t)\Bigg|_{(0,0)}^{(\mathrm{Kx}-1,\mathrm{Ky}-1)} =$$

$$\sum_{m=\mathrm{r}}^{M+r-\mathrm{Kx}} \sum_{n=\mathrm{t}}^{N+t-\mathrm{Ky}} \left( \begin{array}{c} \frac{\partial E}{\partial x_{\mathrm{i}}^{l+1}(m-r,n-r)} \times \frac{\partial x_{\mathrm{i}}^{l+1}(m-r,n-t)}{\partial P_i^{l+1}[\Psi_i^{l+1}(y_k^l(m-r,n-t),w_{ik}^{l+1}(0{,}0)),..,\Psi_i^{l+1}(y_k^l(m,n),w_{ik}^{l+1}(\mathrm{r,t}),)\dots)]} \times \\ \frac{\partial P_i^{l+1}\left[\Psi_i^{l+1}\left(y_k^l(m-r,n-t),w_{ik}^{l+1}(0{,}0)\right),..,\Psi_i^{l+1}(y_k^l(m,n),w_{ik}^{l+1}(\mathrm{r,t}),)\dots)\right]}{\partial \Psi_{ik}^{l+1}\left(y_k^l(m,n),w_{ik}^{l+1}(r,t)\right)} \times \\ \frac{\partial \Psi_{ik}^{l+1}(y_k^l(m,n),w_{ik}^{l+1}(r,t))}{\partial w_{ik}^{l+1}(r,t)} \end{array} \right) \tag{32}$$

*where* $\frac{\partial x_{\mathrm{i}}^{l+1}(m-r,n-t)}{\partial P_i^{l+1}[\Psi_i^{l+1}(y_k^l(m-r,n-t),w_{ik}^{l+1}(0{,}0)),..,\Psi_i^{l+1}(y_k^l(m,n),w_{ik}^{l+1}(\mathrm{r,t}),)\dots)]} = 1.$

Let $\nabla_w \Psi_{ki}^{l+1}(m,n,r,t) = \frac{\partial \Psi_{ik}^{l+1}\left(y_k^l(m,n),w_{ik}^{l+1}(r,t)\right)}{\partial w_{ik}^{l+1}(r,t)}$, then it simplifies to:

$$\frac{\partial E}{\partial w_{ik}^{l+1}}(r,t)\Bigg|_{(0,0)}^{(2,2)} = \sum_{m_0=\mathrm{r}}^{M+r-1} \sum_{n_0=\mathrm{t}}^{N+t-1} \Delta_{\mathrm{i}}^{l+1}(m_0-r,n_0-t) \times \nabla_{\Psi_{ki}} \mathrm{P}_i^{l+1}(m_0,n_0,r,t) \times \nabla_w \Psi_{ki}^{l+1}(m_0,n_0,r,t)$$

$$\textit{Let } \nabla_{\mathrm{w}} \mathrm{P}_i^{l+1}(m_0,n_0,r,t) = \nabla_{\Psi_{ki}} \mathrm{P}_i^{l+1}(m_0,n_0,r,t) \times \nabla_w \Psi_{ki}^{l+1}(m_0,n_0,r,t),$$

$$\frac{\partial E}{\partial w_{ik}^{l+1}}(r,t)\Bigg|_{(0,0)}^{(2,2)} = \sum_{m_0=\mathrm{r}}^{M+r-1} \sum_{n_0=\mathrm{t}}^{N+t-1} \Delta_{\mathrm{i}}^{l+1}(m_0-r,m_0-t) \times \nabla_{\mathrm{w}} \mathrm{P}_i^{l+1}(m_0,n_0,r,t), \qquad \textit{or let } m = m_0 - r, n = n_0 - t \tag{33}$$

$$\frac{\partial E}{\partial w_{ik}^{l+1}}(r,t)\Bigg|_{(0,0)}^{(2,2)} = \sum_{m=0}^{M-1} \sum_{n=0}^{N-1} \Delta_{\mathrm{i}}^{l+1}(m,n) \times \nabla_{\mathrm{w}} \mathrm{P}_i^{l+1}(m+r,n+t,r,t)$$

$$\therefore \frac{\partial E}{\partial w_{ik}^{l+1}} = Conv2Dvar(\Delta_{\mathrm{i}}^{l+1}, \nabla_{\mathrm{w}} \mathrm{P}_i^{l+1})$$

Note that the first term, $\Delta_{\mathrm{i}}^{l+1}(m,n)$, in Eq. (33) is a 2D map (matrix) independent of the kernel indices, *r* and *t*. It will be element-wise multiplied by the other two latter terms, each with the same dimension, (i.e., *M-2xN-2* for $K_x$=$K_y$=3) and created by derivative functions of nodal and pool operators applied over the pixels of the *MxN* output, $y_k^l$, and the corresponding weight value, $w_{ik}^{l+1}(r,t)$. Note that although $w_{ik}^{l+1}(r,t)$; is fixed for each shift value, *r* and *t*; the pixels $y_k^l(m,n)$ are taken from different (shifted) sections of $y_k^l$. This operation is illustrated in Figure 15.

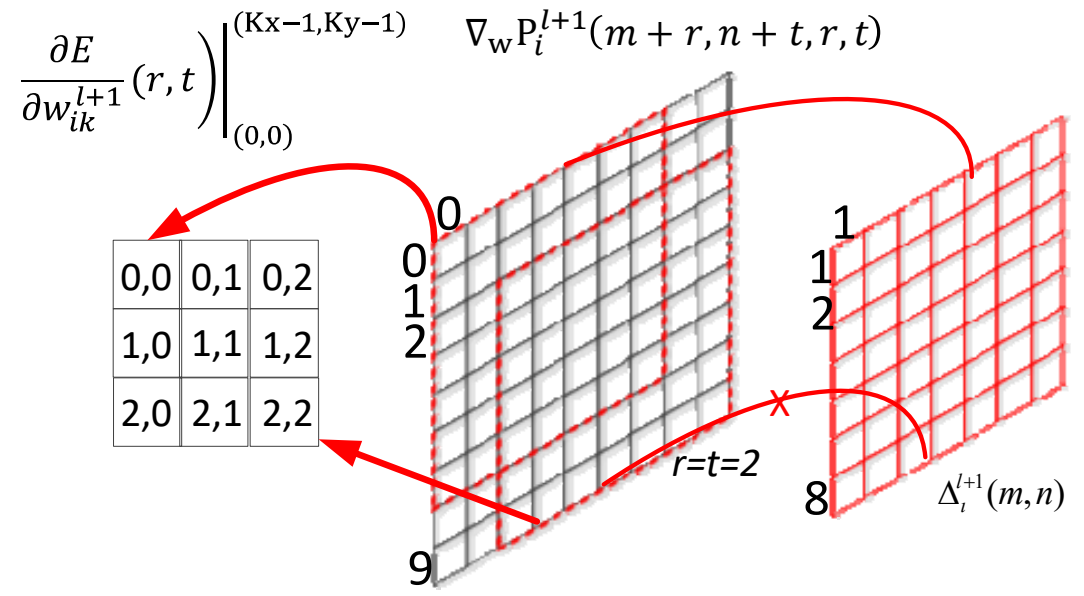

**Figure 15: Computation of the kernel sensitivities.**

To accommodate the boundary conditions for Eq. (33), once again consider the simple ONN in Figure 16. The generic weight sensitivity expression in Eq. (33) can accommodate the boundary conditions by properly forming the 4D matrices, $\nabla_{\Psi_{ki}} \mathrm{P}_i^{l+1}$ and $\nabla_w \Psi_{ki}^{l+1}$. The former 4D matrix is already formed properly by considering the boundary pixels of the output map, $y_k^l(m,n)$. Similarly, the latter 4D matrix, $\nabla_w \Psi_{ki}^{l+1}$, is now formed by considering the filter parameters, $w_{ik}^{l+1}(r,t)$ that operate with the output map pixels. For the toy ONN, $\nabla_{\mathrm{w}} \mathrm{P}_i^{l+1}(m,n,r,t) = y_k^l(m,n)$ whenever applicable, and Figure 16 illustrates the pixels of $y_0^1$ that are operated with the weight elements, $w_0^2(0{,}0)$, and $w_0^2(1{,}1)$ of the output neuron. It is obvious that the boundary pixels of $y_0^1$ do not operate with *all* the weight elements and naturally this will yield $\nabla_{\Psi_{ki}} \mathrm{P}_0^2(m,n,r,t) = 0$ for those pixel elements, $y_0^1(m,n)$ that do not operate with $w_0^2(r,t)$.

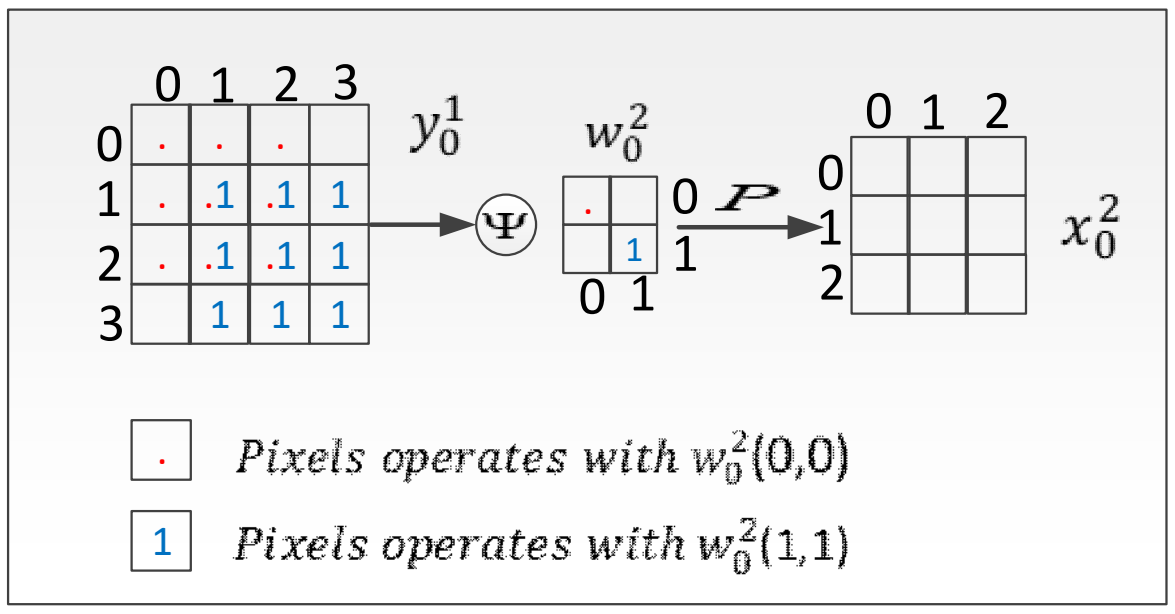

**Figure 16: The pixels of $y_0^1(m,n)$ that are operated with the weight elements of the output neuron, $w_0^2(0,0)$, and $w_0^2(1,1)$.**

Further implementation details for BP training are in [37].